\documentclass[onefignum,onetabnum]{siamonline190516}

\usepackage{lipsum}
\usepackage{amsfonts}
\usepackage{graphicx}
\usepackage{epstopdf}
\usepackage{algorithmic}
\ifpdf
  \DeclareGraphicsExtensions{.eps,.pdf,.png,.jpg}
\else
  \DeclareGraphicsExtensions{.eps}
\fi

\usepackage{enumitem}
\setlist[enumerate]{leftmargin=.5in}
\setlist[itemize]{leftmargin=.5in}

\newsiamremark{remark}{Remark}
\newsiamremark{hypothesis}{Hypothesis}
\crefname{hypothesis}{Hypothesis}{Hypotheses}
\newsiamthm{claim}{Claim}

\headers{Tensor-based framework for training flexible neural networks}{Y. Zniyed, K. Usevich, S. Miron, and D. Brie}

\title{Tensor-based framework for training flexible neural networks\thanks{Submitted to the editors DATE.
\funding{This research was supported by the ANR (Agence Nationale de Recherche) grant LeaFleT (ANR-19-CE23-0021).}}}

\author{Yassine Zniyed\thanks{CRAN, Universit\'e de Lorraine, CNRS, Vand\oe{}uvre-l\`es-Nancy, France 
  (\email{firstname.lastname@univ-lorraine.fr}).}
\and Konstantin Usevich\footnotemark[2]
\and Sebastian Miron\footnotemark[2]
\and David Brie\footnotemark[2]}

\usepackage{amsopn}

\usepackage{amssymb,mathrsfs}

\definecolor{darkgreen}{rgb}{0,0.46,0}

\newcommand{\vect}[1]{\boldsymbol{#1}}

\newcommand{\set}[1]{\mathscr{#1}}

\ifpdf
\hypersetup{
  pdftitle={Tensor-based framework for training flexible neural networks},
  pdfauthor={Y. Zniyed, K. Usevich, S. Miron, and D. Brie}
}
\fi

\begin{document}

\maketitle

\begin{abstract}
Activation functions (AFs) are an important part of the design of neural networks (NNs), and their choice plays a predominant role in the performance of a NN. In this work, we are particularly interested in the estimation of flexible activation functions using tensor-based solutions, where the AFs are expressed as a weighted sum of predefined basis functions.
To do so, we propose a new learning algorithm which solves a constrained coupled matrix-tensor factorization (CMTF) problem. This technique fuses the first and zeroth order information of the NN, where the first-order information is contained in a Jacobian tensor, following a constrained canonical polyadic decomposition (CPD).  
The proposed algorithm can handle different decomposition bases. The goal of this method is to compress large pretrained NN models, by replacing subnetworks, {\em i.e.,} one or multiple layers of the original network, by a new flexible layer. The approach is applied to a pretrained convolutional neural network (CNN) used for character classification. 
\end{abstract}



\section{Introduction}
Neural networks (NNs) are powerful tools, providing solutions to a wide range of real world problems \cite{lecun2015deeplearning}. The reason behind their omnipresence is their power to learn to mimic, and to approximate complex nonlinear functions, while showing good prediction performances in numerous applications. Training a NN requires making a number of decisions, such as the number of hidden layers and nodes, and the shape of the activation functions (AFs). The choice of AFs is a critical step for training a NN, and their choice plays a predominant role in the performance of the network.

From a theoretical point of view, the universal approximation theorem (UAT) \cite{HORNIK1989359} states that a neural network with one, wide enough, hidden layer can approximate any continuous function for inputs within a specific range, under mild conditions on the AFs. These conditions are that the AFs must be non-constant, bounded, monotonically-increasing and continuous. This implies that we can approximate any function by only increasing the width of a single layer. In the neural network literature, the AFs are usually chosen as fixed-shape functions, such as sigmoid \cite{cybenko1988continuous}, hyperbolic tangent \cite{55123} or ReLU \cite{pmlr-v15-glorot11a}, to name a few. The use of this type of functions may face certain limitations. We cite for example the vanishing gradient problem \cite{279181} for the bounded AFs, and the dying ReLU problem \cite{Maas13rectifiernonlinearities} for the ReLU function.

In order to improve the performances of NNs, an up-and-coming approach, called flexible activation functions (FAFs), also known as trainable or learnable activation functions, was considered. It is an ambitious approach, which allows to find the FAFs by learning. In this area, the AFs can be classified into two categories. The first one is the parameterized standard AFs, where the AFs are based on the fixed-shape AFs with one or more learnable parameters. In this category, we can find sigmoid-like functions, such as the generalized sigmoid \cite{NARAYAN199769} and the sigmoidal selector \cite{CHANDRA2004429}, or other ReLU-like functions, such as the Parametric ReLU \cite{10.1109/ICCV.2015.123}, the Flexible ReLU \cite{8546022} or the Swish \cite{Ramachandran2018}. The use of these parameterized AFs is useful from a practical point of view, since they are trained from the data, and tend to converge more quickly than the fixed AFs \cite{8913972}. However, these functions have shown a poor increase of the networks expressiveness, compared to their standard fixed-shape versions \cite{APICELLA202114}.

A second category of FAFs concerns the functions based on ensemble methods. We refer the reader to \cite{APICELLA202114}, for a recent and a complete survey on this subject. The approaches in this category propose to combine different basis functions or models, together, to approximate a general FAF. In this work, we define a new NN model having different and flexible AFs in each neuron. The proposed flexible AFs are expressed as a weighted sum of predefined bases elements. 
Different bases are considered in this work. This NN will be referred to in the sequel by the flexible NN. The interest of such a model is to approximate the input-output relationship of nonlinear systems, using a low number of neurons, by increasing the expressiveness of the network.
Examples of FAFs are \cite{Goyal2019LearningAF,Livni2013AnAF,SHI201687}, where the polynomial basis is used to learn a general multivariate polynomial function, \cite{Lopez2019}, where the AF is expressed as a piecewise polynomial function with a different AF for each neuron, but the solution is limited to the one-output one-hidden layer models, and \cite{agostinelli2015learning}, where each neuron is characterized by his independent piecewise linear AF. The various works on this subject have shown an improvement in performances compared to the fixed-shape functions \cite{APICELLA202114}.
However, the research in this category is still far from being mature. The FAFs-related works often propose different learning algorithms for each set of basis functions, {\em i.e.,} each algorithm has been developed for a specific basis. Moreover, the research in the design of the activation functions domain is active, but the proposed learning processes are based on conventional training techniques.

In this paper, we propose a tensor-based methodology, borrowed from the system identification community \cite{Dreesen2015}, to train the flexible NN models. It should be noted that NNs can be viewed as tools for decomposing multivariate functions into sums and compositions of univariate functions. An interesting approach to decompose multivariate functions is the Jacobian tensor decomposition. This method was first developed in \cite{Dreesen2015}, where the authors have proposed to use a $3$-order tensor built from Jacobian matrices of a multivariate polynomial function. In \cite{Dreesen2015}, the method allows to decompose the multivariate polynomial function as linear combinations of univariate polynomials in linear forms of the input variables. This method is based on an unstrucutred canonical polyadic decomposition (CPD) \cite{Hitchcock1927, Harshman1970}, which is critical. In practice, we need to constrain the decomposition, for two reasons. First, to be able to recover meaningful univariate nonlinearities, and second, to improve the identifiability of the model \cite{KU_XRANK}. Other works \cite{Hollander2018, Zniyed2021} have proposed to add constraints on the model. However, all these methods $(i)$ consider the decomposition of multivariate polynomial functions only, $(ii)$ rely on the Jacobian decomposition and thus can not estimate the constant terms of the approximated functions, which will introduce a bias in the approximation, and $(iii)$ can not be applied directly to NNs, as will be shown later.

In this work, we develop a framework for compressing large neural networks. The aim of this work is to replace pretrained subnetworks, {\em i.e.,} one or multiple layers of the original network, with one new flexible layer, having different and flexible AFs, with the purpose of reducing the number of parameters \cite{doi:10.1137/19M1296070, doi:10.1137/19M1246468}. This method can be applied to different subnetworks of the original network. In order to retrieve the FAFs, two major propositions are developed. First, we formulate the learning problem as a constrained tensor decomposition (CTD) problem, and propose improvement to the state-of-the-art solutions to have a more computationally efficient solution. Second, we propose to fuse the first and zeroth order information, by reformulating the learning problem as a constrained coupled matrix-tensor factorization (CMTF) \cite{Acar2011AllatonceOF} problem. The matrix in this problem corresponds to the output of the original subnetwork, and the tensor to the first-order derivatives of the function between the input and the output of this subnetwork. This reformulation allows to overcome the CTD formulation drawbacks. The proposed learning algorithm is a constrained alternating least squares (ALS) algorithm that can handle different basis functions, such as, but not limited to, piecewise linear functions, piecewise polynomial functions and polynomial functions. This solution applies nonlinear constraints on the factors of the CMTF. We will show that the proposed CMTF formulation allows to improve the functions approximation, compared to the constrained decomposition of the Jacobian tensor, even when the constant terms are estimated in this latter.

The contributions of this work can be summarized as follows.
\begin{enumerate}
    \item We consider a neural network model where each neuron has a different and flexible AF. The FAFs are expressed as a weighted sum of predefined basis functions.
    \item As a first step, we formulate the learning problem as a constrained tensor decomposition problem. We propose a computationally more efficient solution to this problem.
    \item To overcome the CTD-based learning drawbacks, we reformulate the problem as a constrained coupled matrix-tensor factorization problem, combining the first and zeroth order information. We propose an ALS-based solution to this problem and show the improvements of performances with this strategy.
    \item As opposed to the the state-of-the-art solutions, the proposed approaches can handle different basis functions.
    \item Some suggestions and explanations are given to improve the implementation of the learning algorithm, speed up its execution and to build the Jacobian tensor from the original subnetwork using automatic differentiation (AD) \cite{Baydin2017} techniques.
\end{enumerate}

The rest of the paper is organized as follows. 
Section \ref{sec3} defines the proposed flexible model. Examples of flexible AFs are given in this part.
In Section \ref{sec4}, we present the tensor-based approach for decomposing multivariate polynomial functions. Related works and limitations of this approach are discussed. An overview on tensor decompositions, in general, and the Jacobian tensor decomposition, in particular, are also given.
Section \ref{sec5} formulates the tensor-based learning problem as an optimization problem. A new solution to solve the CTD-based problem is proposed in this section. The limitations of the CTD-based approaches are pointed out.
Section \ref{sec6} details the new proposed CMTF formulation and presents the proposed learning algorithm. An illustrative example of multivariate function decomposition, considering a toy example, is given in this section.
Section \ref{sec7} is dedicated to the neural networks compression. 
This section shows how to apply the proposed flexible model to pretrained neural networks. Furthermore, it shows the effectiveness of the proposed solution through numerical simulations.
Section \ref{sec8} draws the conclusions and perspectives.

\section{Flexible activation function model}
\label{sec3}
\subsection{Flexible model}
\begin{figure}[htbp]
\centering\includegraphics[scale=0.05]{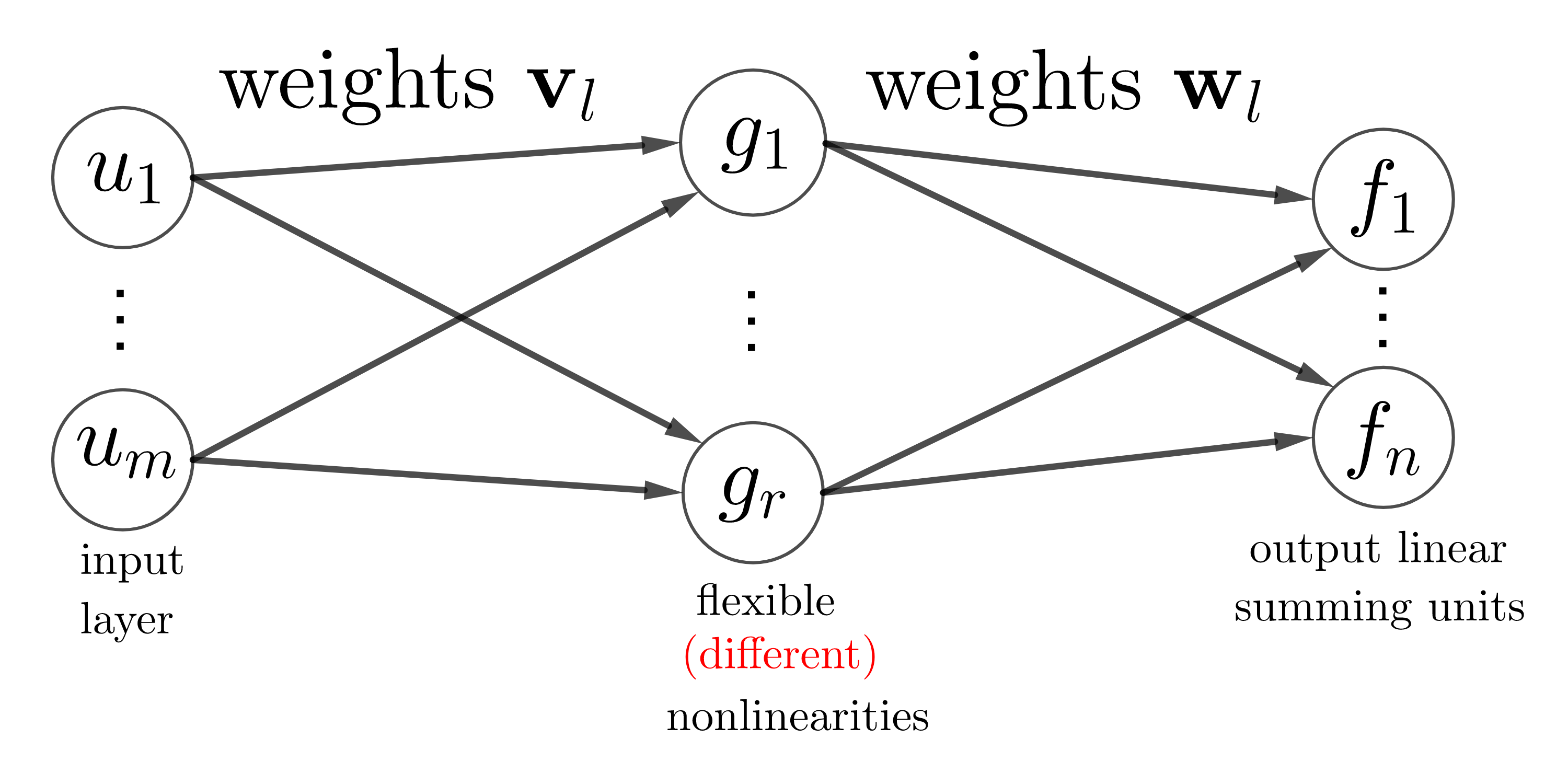}
\caption{Graphical representation of the flexible NN.}
\label{fig_fun}
\end{figure}

The model proposed in this work is a flexible NN model, where each neuron has a different and flexible AF. A graphical illustration of the basic one-layer block of this model is given in Fig. \ref{fig_fun}.

It is known in the literature that the Kolmogorov–Arnold representation theorem \cite{Kolmogorov, Arnold, 6797106} states that any continuous function $f$ can be representable by sums and superpositions of continuous univariate functions. This result is very important in the context of neural networks, since the goal is the approximation of multivariate functions by such representations.
The model proposed in this work is quite similar to Kolmogorov's representation, in the sense that his representation has different univariate functions.
Following this idea, the rationale behind the proposed model is to increase the expressiveness of the one-layer network using different and flexible activation functions. This will allow to approximate complex nonlinear functions with a lower number of neurons compared to classical activation functions.

Formally, the multivariate function that represents the input-output relationship is denoted by $\mathbf{f}$. This function $\mathbf{f}: \mathbb{R}^m \rightarrow \mathbb{R}^n$ is vector-valued, and is defined as:
\begin{align}
\mathbf{f}(\mathbf{u}) = [{f}_1(\mathbf{u}) \cdots {f}_n(\mathbf{u})]^T,\\
\mbox{with ~\ ~\ ~\ } \mathbf{u} = [u_1 \cdots u_m]^T.
\end{align}
The function $\mathbf{f}$ is expressed as:
\begin{align}
\label{eq_problem}
\mathbf{f}(\mathbf{u}) = \mathbf{W} \mathbf{g}(\mathbf{V}^T \mathbf{u}),
\end{align}
where $\mathbf{V} \in \mathbb{R}^{m \times r}$, $\mathbf{W} \in \mathbb{R}^{n \times r}$ are transformation matrices, $\mathbf{v}_l$ and $\mathbf{w}_l$, for $1 \leq  l \leq  r$, are respectively the columns of $\mathbf{V}$ and $\mathbf{W}$, and $\mathbf{g}: \mathbb{R}^r \rightarrow \mathbb{R}^r$ follows
\begin{align}
\label{eq_AF}
\mathbf{g}(t_1, \cdots, t_r) = [{g}_1(t_1) \cdots {g}_r(t_r)]^T,
\end{align}
with $g_l: \mathbb{R} \rightarrow \mathbb{R}$ is a univariate nonlinear function.
Equivalently, the function in \eqref{eq_problem} can be expressed as
\begin{align}
\label{eq_problem_equiv}
\mathbf{f}(\mathbf{u}) = \mathbf{w}_1 \cdot g_1(\mathbf{v}_1^T \mathbf{u}) + \cdots + \mathbf{w}_r \cdot g_r(\mathbf{v}_r^T \mathbf{u}).
\end{align}

\subsection{Basis functions}
The univariate nonlinear functions $g_l$ ($1 \leq  l \leq  r$) in \eqref{eq_problem_equiv} are represented as a linear combination of predefined basis functions $\{\phi_1, \hdots, \phi_d\}$.  The function $g_l$ is then expressed as:
\begin{align}
\label{model_AF}
g_l(t)= c_{0,l} + c_{1,l} \phi_1(t) + \cdots + c_{d,l} \phi_d(t).
\end{align}
The functions $\phi_k$, for $1 \leq  k \leq  d$, in \eqref{model_AF} are \textit{a priori} chosen functions, and the parameters $c_{k,l}$, for $0 \leq  k \leq  d$ and $1 \leq  l \leq  r$, are the learnable parameters.

\begin{figure}[htbp]
\centering\includegraphics[width=\linewidth,height=10cm]{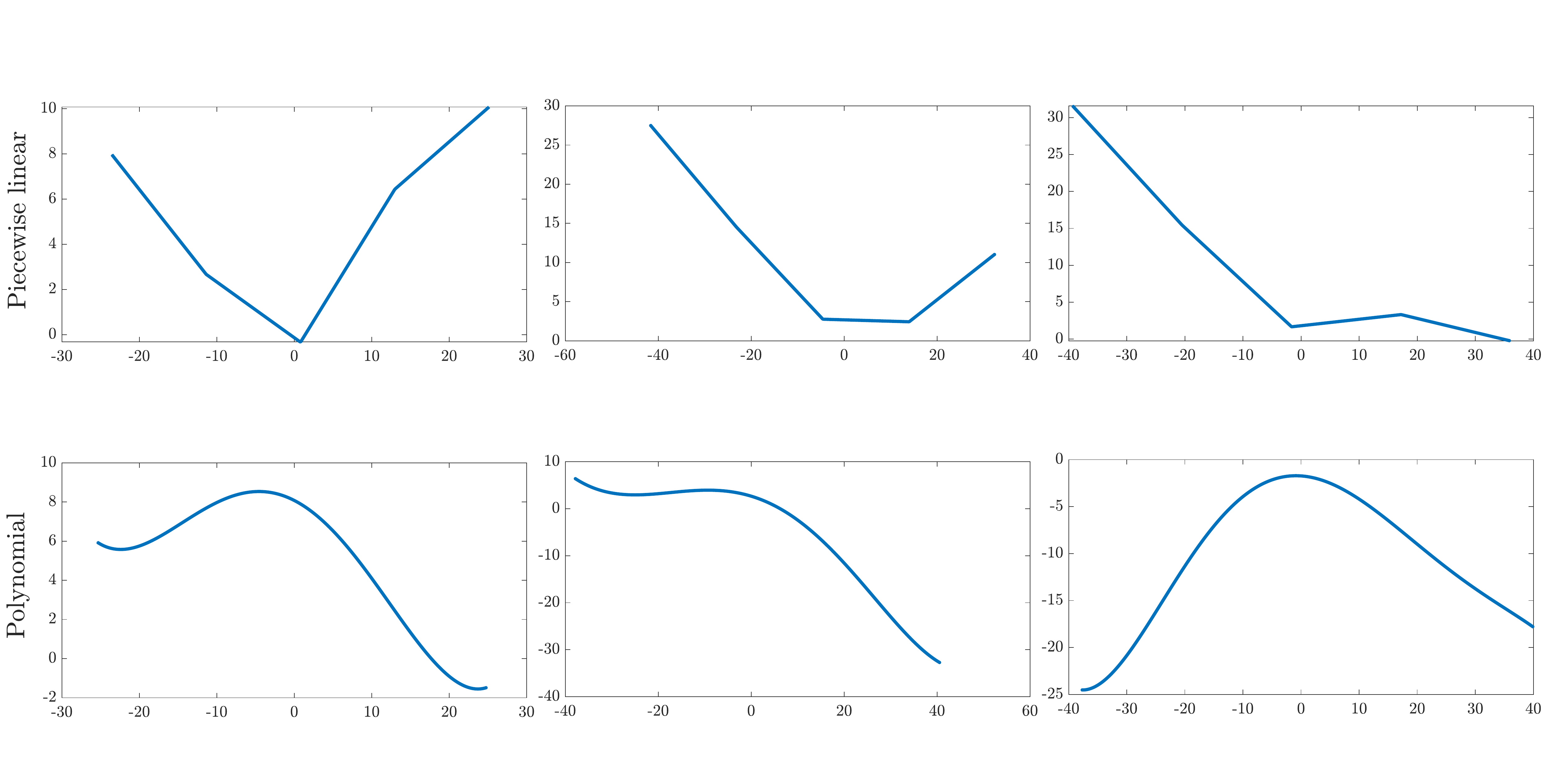}
\caption{Examples of flexible AFs}
\label{fig_ex_FAFs}
\end{figure}

We give, in Fig. \ref{fig_ex_FAFs}, some examples of the flexible AFs, obtained using the learning algorithm (Algorithm \ref{alg_learning}) which will be presented later.
In this figure, we present, without loss of generality, AFs which are piecewise linear functions $\big(\phi_k(t) = {\rm ReLU}(t-t_k)\big)$ and polynomial functions $\big(\phi_k(t) = t^k\big)$. This shows that the FAFs can have several different shapes depending on the decomposition basis, as shown in the figure. Other more or less simple forms of the FAFs have also been found, but are not presented in this figure.

\subsection{Polynomial bases and X-rank decomposition}
In the case when the functions are polynomial, i.e., the basis functions are chosen as $\phi_k(t) = t^{k}$, the decomposition \eqref{eq_problem_equiv} is a special case of the so-called X-rank decomposition \cite{KU_XRANK}.
In particular, it can be shown that the flexible neural network representation \eqref{eq_problem_equiv} is, under certain conditions, unique, which may be helpful for ensuring stability and interpretability of the representation.

Formally, the X-rank decomposition is defined as follows. For a vector space $\set{A}$ and a set of rank-one terms  $\set{X} \subset \set{A}$, the X-rank of a vector $\vect{v}$ is defined as the smallest number of rank-one elements, such that $\vect{v}$ can be represented as their sum: 
\begin{equation}\label{eq:xrank}
\operatorname{rank}_{\set{X}}{\vect{v} } = \min r: \vect{v}  = \vect{x}_1 + \cdots +\vect{x}_r, \quad \vect{x}_l \in {\set{X}}.
\end{equation}
Decomposition \eqref{eq:xrank} in general is a so-called atomic decomposition.
In order for  \eqref{eq:xrank} to be an X-rank decomposition,  the set $\set{X}$ has to satisfy certain properties \cite{KU_XRANK} (i.e. $\set{X}$  has to be an irreducible non-degenerate algebraic variety).
In the case of polynomial AFs, once we fix the degree of polynomials, the map $\vect{f}(\cdot)$ belongs to the vector space of polynomial functions  $\mathbb{R}^{m} \to \mathbb{R}^{n}$ of degree $d$, and the set of ``rank-one'' terms is the set of functions $\vect{w} g(\vect{v}^{T} \vect{u})$.

It is worth mentioning that the decomposition \eqref{eq_problem_equiv} is associated with some strong results on the identifiability, when the AFs are polynomials.
Indeed, it has been shown in \cite{KU_XRANK}, that for the case where $d \leq 3$ and $m \leq 2$, the decomposition in \eqref{eq_problem_equiv} is unique if 
\begin{equation}
r \leq \min (m, \left \lceil r_5(m,n,d) \right \rceil - 1) \cdot n,
\end{equation}
where $r_5(m,n,d)=\frac{\binom{m+d-1}{d}}{m+n-1}$ except some tuples $(m,n,d)$ \cite{KU_XRANK}.
In the same case where $d \leq 3$, $m \leq 2$ and $r > mn$, the authors have proved that the model \eqref{eq_problem_equiv} is not identifiable.

Another interesting result associated with the X-rank decomposition of polynomial functions is the partial identiafibility, when $r > mn$. This result \cite{KU_XRANK} states that for $1 \leq s \leq d$, and for all $r \leq \binom{m+s-1}{s} \cdot n$, the decomposition \eqref{eq_problem_equiv} is (partially) identifiable, except the coefficients $c_{k,l}$, for $k < s$.

These interesting results show that in the case of polynomial AFs, the decomposition \eqref{eq_problem_equiv} can be unique/identifiable even if the associated tensor decomposition problem, that we will detail in the next section, is not unique. The results for other decomposition bases remain open problems, but we believe that these kind of results can be useful to prove and guarantee the stability and interpretability of neural networks.

\section{Tensor-based multivariate function decomposition}
\label{sec4}
In this section, we first present some motivations for using tensor decompositions. Some useful notations and definitions are presented. We then recall the tensor-based method for decomposing multivariate functions. We will present in particular the method in \cite{Dreesen2015}, where the authors use a Jacobian tensor of a multivariate polynomial function. Then, we will outline the works which have proposed to add polynomial constraints to the decomposition, in the context of system identification, and finally we will discuss the limitations of all these methods for the neural networks learning.


\subsection{What are tensor decompositions}
Tensor decompositions \cite{Kolda2009, comon2014} are powerful tools of multilinear algebra, which have been used in a variety of applications. Tensors are natural generalization of matrices ($2$-order tensors), which enable to model data with more than two axes of variation. They are characterized by the order, which is the number of their dimensions. Tensors quickly became an important tool in several applications, including data mining and machine learning \cite{Sidi2017}.
It is worth noting that tensors have been used in the context of neural networks, we cite among these works \cite{pmlr-v49-cohen16}, where the authors focused on relating tensor decompositions to neural networks with product units, instead of summing units. However, this architecture is not so much used in practice. Other tensor-based approaches \cite{Lebedev2015SpeedingupCN,kim2016compression, 10.5555/2969239.2969289} have been proposed for the compression of neural networks. The aim of these works was to compress neural networks by finding low-rank approximations of the weight tensors.

The reasons for this interest in tensors are mainly three. First, being a natural tool that has the ability to model heterogeneous data. Second, having interesting uniqueness properties, which is a real advantage over matrices, and finally the availability of tools to perform tensor decompositions. Different tensor decompositions exist in the literature, we mention among them, the canonical polyadic decomposition \cite{Hitchcock1927, Harshman1970}. The CPD decomposes a tensor $\boldsymbol{\mathcal{X}}$ into a sum of $r$ rank-one tensors. A rank-one tensor of order $Q$ is given by the outer product of $Q$ vectors. The canonical rank $r$ of a CPD is the minimum number of rank-one tensors needed for a perfect representation of $\boldsymbol{\mathcal{X}}$. A more formal definition of the CPD is given below.

\paragraph{Notations and definitions} The notations used throughout the rest of this paper are now defined. The symbols $(\cdot)^{\dagger} $ and ${\rm rank}(\cdot)$ denote, respectively, the pseudo-inverse and the rank. The outer, Kronecker, Khatri-Rao, Hadamard and $n$-mode products are denoted to by $\circ$, $\otimes$, $\odot$, $*$ and $\times_n$. Tensors are represented by $\boldsymbol{\mathcal{X}}$. $\boldsymbol{\mathcal{X}}(i,:,:)$, $\boldsymbol{\mathcal{X}}(:,j,:)$ and $\boldsymbol{\mathcal{X}}(:,:,k)$ are the $i$-th horizontal, $j$-th lateral and $k$-th frontal slices of sizes $N_2 \times, N_3$, $N_1 \times N_3$ and $N_1 \times N_2$, respectively, of the tensor $\boldsymbol{\mathcal{X}}$ of size $N_1 \times N_2 \times N_3$. The operator ${\rm vec}(\cdot)$ forms a vector by stacking the columns of its matrix argument. The operator ${\rm diag}(\cdot)$ forms a diagonal matrix from its vector argument. ${\rm unfold}_q \boldsymbol{\mathcal{X}}$ refers to the unfolding of tensor $\boldsymbol{\mathcal{X}}$ over its $q$-th mode \cite{Kolda2009}. We now introduce some definitions that will be useful in the sequel.


\begin{definition}
A $Q$-order tensor $\boldsymbol{\mathcal{X}}$ of size $N_1\times \hdots \times N_Q$ belonging to the family of rank-$r$ CPD can be expressed as:
\begin{align}\label{def_CPD}
\boldsymbol{\mathcal{X}} = \sum_{l=1}^{r}{\mathbf P}_1(:,l) \circ \hdots \circ  {\mathbf P}_Q(:,l),
\end{align}
where the $q$-th factor ${\mathbf P}_q$ is of size $N_q \times r$, for $1\leq q \leq Q$. An equivalent notation to denote the CP decomposition in \eqref{def_CPD} is $\boldsymbol{\mathcal{X}} = [| {\mathbf P}_1,\cdots,{\mathbf P}_Q |]$.
The $q$-mode unfolded matrix ${\rm unfold}_q \boldsymbol{\mathcal{X}}$, of size $N_q \times \frac{N_1 \cdots N_Q}{N_q}$, is given by:
\begin{align}
{\rm unfold}_q \boldsymbol{\mathcal{X}} = \mathbf{P}_q \cdot (\mathbf{P}_Q \odot \cdots \odot \mathbf{P}_{q+1} \odot \mathbf{P}_{q-1} \odot \cdots \odot \mathbf{P}_1)^T.
\end{align}
\end{definition}

The CP decomposition in \eqref{def_CPD} is said to be unique (up to trivial scaling and permutation ambiguities) if 
\begin{equation}
\label{eq_kruskal}
    \sum_{q=1}^{Q} k_{\mathbf{P}_q} \leq 2 r + (Q-1),
\end{equation}
where $k_{\mathbf{P}_q}$ is the Kruskal rank \cite{Kolda2009} of factor $\mathbf{P}_q$. The condition in \eqref{eq_kruskal} is a sufficient condition. It has been proven in \cite{KRUSKAL197795} of $3$-order CPDs, and extended to high-order CPDs in \cite{Sidi_Bro2000}.
\subsection{The Jacobian tensor method}
\label{sec_tensor_construction}
Several authors \cite{Schoukens2012, Mulders2014, Dreesen2015} have been focused in the representation of a given complex nonlinear function $\mathbf{f}$ by a simpler representation, such as \eqref{eq_problem}. This representation allows to approximate the function, reduce the number of its parameters, or give a better understanding of the mechanisms of nonlinear mappings. These methods assume that the function $\mathbf{f}$ is known through a polynomial parametric representation. 
These methods were developed by the nonlinear system identification community, in the context of the decoupling problem for multivariate real polynomials.
One of these methods is the Jacobian method \cite{Dreesen2015}. This method tries to decompose $\mathbf{f}(\mathbf{u})$ according to \eqref{eq_problem} relying on its first-order information. To do so, the Jacobian method considers the decomposition of a single $3$-order tensor, unlike the coefficient-based methods \cite{Schoukens2012},  {\em i.e.,} methods that use directly the coefficients of $\mathbf{f}(\mathbf{u})$, which require the decomposition of several high-order tensor for high degree polynomials.

\begin{figure}[htbp]
\centering
\includegraphics[scale=0.2]{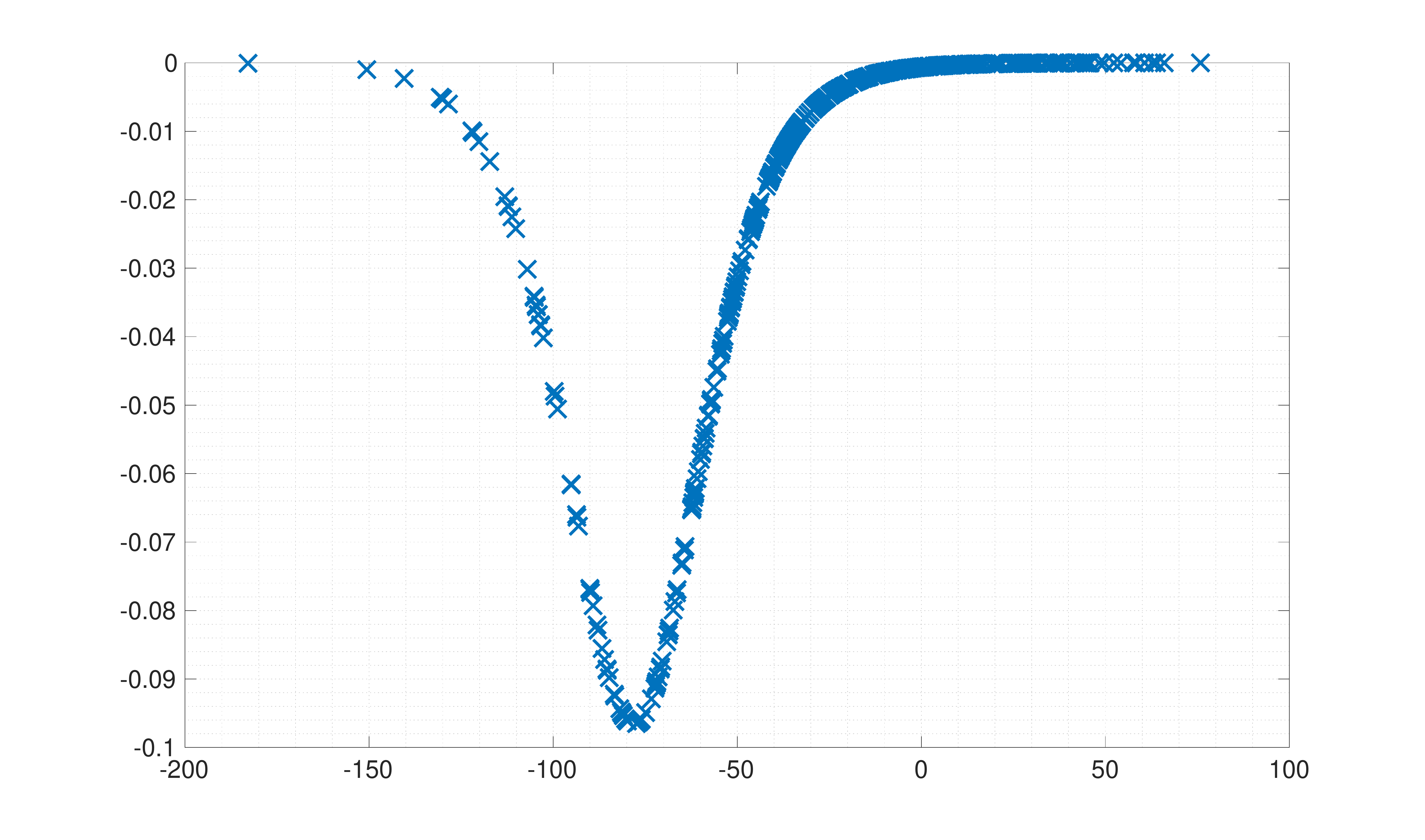}
\caption{$3$-th column of $\mathbf{H}$ against projected points $(\mathbf{v}_3^T\mathbf{u}^{(1)}, \cdots, \mathbf{v}_3^T\mathbf{u}^{(N)})$.}
\label{fig1}
\end{figure}

The Jacobian method relies on the observation that the Jacobian matrix of $\mathbf{f}$ in point $\mathbf{u}$ is expressed as
\begin{align}
\mathbf{J}_{\mathbf{f}} (\mathbf{u}) := \begin{bmatrix}
\frac{\partial f_1}{\partial u_1} (\mathbf{u})& \cdots & \frac{\partial f_1}{\partial u_m} (\mathbf{u}) \\ 
\vdots &  & \vdots \\ 
\frac{\partial f_n}{\partial u_1} (\mathbf{u}) & \cdots & \frac{\partial f_n}{\partial u_m} (\mathbf{u})
\end{bmatrix}= \mathbf{W} \cdot diag\big( g_1'(\mathbf{v}_1^T \mathbf{u}) \cdots g_r'(\mathbf{v}_r^T \mathbf{u})\big) \cdot \mathbf{V}^T.
\end{align}
This result follows by applying the chain rule \cite{Dreesen2015}. By stacking the Jacobian evaluations at $N$ different sampling points $\mathbf{u}^{(j)} \in \mathbb{R}^m$, for $1 \leq j \leq N$, such that
\begin{align}
\boldsymbol{\mathcal{J}}_{:,:,j} = \mathbf{J}_{\mathbf{f}} (\mathbf{u}^{(j)}),
\end{align}
we construct a $3$-order tensor $\boldsymbol{\mathcal{J}}$ of size $n \times m \times N$, following a CPD expressed as
\begin{align}
\label{tensor_J}
\boldsymbol{\mathcal{J}} = \sum_{l=1}^{r}{\mathbf w}_l \circ {\mathbf v}_l \circ  {\mathbf h}_l,
\end{align}
or equivalently as,
\begin{align}
\boldsymbol{\mathcal{J}} = [| {\mathbf W}, {\mathbf V}, {\mathbf H} |].
\end{align}
The columns of $\mathbf{H} \in \mathbb{R}^{N \times r}$ have the following structure \cite{Dreesen2015, USEVICH202022}
\begin{align}
\label{eq_structure}
\mathbf{h}_l = [ g_l'(\mathbf{v}_l^T \mathbf{u}^{(1)}), \cdots  , g_l'(\mathbf{v}_l^T \mathbf{u}^{(N)}) ]^T.
\end{align}
The aim of this approach is to determine the coefficients of function $g_l$ from the retrieved CPD factors $\mathbf{V}$, $\mathbf{W}$ and $\mathbf{H}$.
When the decomposition \eqref{tensor_J} is unique, the Jacobian method is a very accurate method for the approximation of AFs. We demonstrate the effectiveness of the Jacobian method in Fig. \ref{fig1}. In this figure, we decompose a unique tensor, of size $10 \times 784 \times 500$ and $r=300$, using a relaxed ALS \cite{Carroll1970, Harshman1970}. This tensor is estimated, using the AD techniques, from a shallow NN with a single hidden layer with sigmoid functions. This NN is trained on the MNIST dataset.
We can see that the smooth curve in Fig. \ref{fig1}, corresponds to the derivative of the considered AF, which shows the effectiveness of this approach for unique tensors. However, when the decomposition is not unique or noisy, the unstructured decomposition \cite{Dreesen2015} will fail. This is discussed in the next section.

\subsection{Related works on the Jacobian decomposition}
Decomposing the tensor $\boldsymbol{\mathcal{J}}$ described in Section \ref{sec_tensor_construction}, has been addressed in various works \cite{Dreesen2015, Hollander2018, Zniyed2021}. In all these works, this decomposition has been proposed in the context of the decoupling problem of multivariate polynomial functions.

First, in \cite{Dreesen2015}, the authors have presented the original idea of the Jacobian tensor, however the decomposition was based on unstructured CPD, {\em i.e.,} without taking into account the structure in \eqref{eq_structure}. This approach is critical, especially when the uniqueness of the CPD is lost. In this case, the algorithm will fail to recover meaningful AFs.

To solve this problem, other works \cite{Hollander2018,Zniyed2021} have proposed to enforce the structure \eqref{eq_structure} on the tensors, in the CP decomposition. In \cite{Hollander2018}, the authors have chosen to parameterize the function $g_l$ using polynomials coefficients, {\em i.e.,} the functions $\{\phi_1, \cdots, \phi_d\}$ in \eqref{model_AF} represent the monomial basis consisting of all monomials of degree $\leq d$. Such solution leads to introduce a polynomial constraint on the third factor of the tensor $\boldsymbol{\mathcal{J}}$. This solution suffers from convergence problems \cite{Zniyed2021} and is not suitable for NN. Indeed, all Jacobian-based methods are not able to estimate the constant terms $c_{0,l}$, for $1 \leq l \leq r$, which will introduce a bias in the approximated function. Besides, it is a computationally intensive solution, which is not adapted for large tensors in the machine learning problems. In \cite{Zniyed2021}, the authors have proposed a reformulation of the problem, which consists in enforcing the structure on the rank-one terms instead of the CPD factors. In the context of NN, this solution suffers from the same problems as \cite{Hollander2018}, namely the constant terms estimation and the heavy computational cost.

In the context of NNs, none of the mentioned works would be effective. This is due to three main drawbacks. Firstly, the original method is based on unstructured decomposition which will fail to recover meaningful AFs in practice. Secondly, the constrained methods have a heavy computational cost and may suffer from convergence problems. Thirdly, all the proposed methods are based on the first-order information, which does not allow to estimate the constant terms of the approximated function.
To overcome these drawbacks, we propose in the next two section, a new formulation of the problem. A first one, where the optimization problem is expressed as a constrained Jacobian tensor decomposition problem, whose solution relies on a new projection strategy, different from the state-of-the-art solutions.
Then, we complete our proposition with a more general CMTF-based formulation of the problem. We will show that this latter solves the drawbacks mentioned before, with a new learning algorithm that is suitable to different types of basis functions and not just polynomials like in \cite{Hollander2018, Zniyed2021}.


\section{Constrained tensor-based approach}
\label{sec5}
This section formulates the CTD-based learning problem as a constrained optimization problem, then a new algorithm to solve this problem is proposed. 
\subsection{CTD problem}
The learning problem, using the Jacobian tensor, can be expressed by the following criterion:
\begin{align}
\label{prob_fac1}
&\min_{\mathbf{W},\mathbf{V},\mathbf{H}} \Big \|  \boldsymbol{\mathcal{J}} - [| \mathbf{W},\mathbf{V},\mathbf{H}|]  \Big \|^2  \\
&\mbox{s.t.}  ~\ ~\ \mathbf{h}_l = \mathbf{X}_l \cdot \mathbf{c}_l, ~\ ~\ ~\ \mbox{for} ~\ 1 \leq l \leq r \notag
\end{align}
where $\mathbf{X}_l$ is of size $N \times (d+1)$, and is expressed as
\begin{align}
\label{matrix_X_l_ctd}
\mathbf{X}_l= \begin{bmatrix}
0 & \phi_1' (\mathbf{v}_l^T \mathbf{u}^{(1)})& \cdots & \phi_d' (\mathbf{v}_l^T \mathbf{u}^{(1)}) \\ 
\vdots & \vdots &  & \vdots \\ 
0 & \phi_1' (\mathbf{v}_l^T \mathbf{u}^{(N)}) & \cdots & \phi_d' (\mathbf{v}_l^T \mathbf{u}^{(N)})
\end{bmatrix}.
\end{align}
Function $\phi_k'(t)$, for $1 \leq k \leq d$, is the derivative of the univariate function $\phi_k(t)$, and the $l$-th coefficient vector $\mathbf{c}_l$ is defined as
\begin{align}
\label{vector_c_l}
\mathbf{c}_l = \begin{bmatrix}
c_{0,l} & c_{1,l} & \hdots & c_{d,l}
\end{bmatrix}^T.
\end{align}
The constraint in \eqref{prob_fac1} means that the entries of matrix $\mathbf{H}$ follow
\begin{align}
\label{const_H}
    h_{j,l}=g_l'(\mathbf{v}_l^T \mathbf{u}^{(j)}),
\end{align}
this allows to satisfy the structural constraint in \eqref{eq_structure}.
The difficulty of the optimization problem in \eqref{prob_fac1} is being a nonlinear and nonconvex optimization problem. In the literature, solutions \cite{Hollander2018, Zniyed2021} have been proposed to solve \eqref{eq_structure}, when the functions are polynomials. These solutions are not computationally efficient.

\subsection{ALS solution}
To solve the problem \eqref{prob_fac1}, we propose an improved constrained alternating least squares (ALS) algorithm, which is computationally more efficient than \cite{Hollander2018}. The idea of this solution is to decompose tensor $\boldsymbol{\mathcal{J}}$ using a relaxed ALS, and then, we project each column $\mathbf{h}_l$ on the column space of $\mathbf{X}_{l}$. In Algorithm \ref{algo_rec2}, we provide the algorithmic description of the ALS-based solution to solve the problem \eqref{prob_fac1}. This solution is composed of two parts, a first part to find an unconstrained decomposition of the tensor $\boldsymbol{\mathcal{J}}$, then a second part to ensure that matrix $\mathbf{H}$ satifies \eqref{const_H}. Vector $\mathbf{c}$, in Algorithm \ref{algo_rec2}, is of length $r(d+1)$, and is defined as $\mathbf{c} = [\mathbf{c_1}; \cdots; \mathbf{c}_r].$
\begin{algorithm}[htbp]
\caption{CTD-based learning algorithm}
\label{algo_rec2}
\begin{flushleft}
\textbf{Input:} Tensor $\boldsymbol{\mathcal{J}}$ of size $n \times m \times N$, functions $\{\phi_1, \cdots, \phi_d\}$, rank $r$\\
\textbf{Output:} Factors $\mathbf{W}$, $\mathbf{V}$ and $\mathbf{H}$, and coefficients $\mathbf{c}_l$.
\end{flushleft}
\begin{algorithmic}[1]
\STATE Initialize $\mathbf{V}$, $\mathbf{H}$
\STATE \textbf{repeat}
~~\\
~~\\
\hrule
~~\\
\STATE{Update $\mathbf{W}$ with $\displaystyle\min_{\mathbf{W}} \Big \| {\rm unfold}_1 \boldsymbol{\mathcal{J}} -  \mathbf{W} \Big( \mathbf{H} \odot \mathbf{V} \Big)^T \Big \|^2$}
\STATE{Update $\mathbf{V}$ with $\displaystyle\min_{\mathbf{V}} \Big \| {\rm unfold}_2 \boldsymbol{\mathcal{J}} -  \mathbf{V} \Big( \mathbf{H} \odot \mathbf{W} \Big)^T \Big \|^2$ }
\STATE{Update $\mathbf{H}$ with $\displaystyle\min_{\mathbf{H}} \left \| {\rm unfold}_3 \boldsymbol{\mathcal{J}} -  \mathbf{H} \Big( \mathbf{V} \odot \mathbf{W} \Big)^T \right \|^2$ }
~~\\
~~\\
\hrule
~~\\
\FOR{$l = 1 \cdots r$}
\STATE{Construct matrix $\mathbf{X}_l$ according to \eqref{matrix_X_l_ctd}}
\ENDFOR
\STATE{Compute $\mathbf{c}$ as $\displaystyle\min_{\mathbf{c}}  \left \| {\rm vec}(\mathbf{H}) - \begin{bmatrix}
\mathbf{X}_1 &  & \text{\large0}\\
 &\ddots & \\
\text{\large0} & & \mathbf{X}_r
\end{bmatrix} \cdot \mathbf{c} \right \|^2$}
\FOR{$l = 1 \cdots r$}
\STATE{Compute $\mathbf{h}_l$ such that: ~\ ~\ $\mathbf{h}_l = \mathbf{X}_l \cdot \mathbf{c}_l$}
\ENDFOR
~~\\
~~\\
\hrule
~~\\
\STATE \textbf{until} a convergence test is satisfied or maximum iterations exhausted
~~\\
\nonumber \hrule
~~\\
\RETURN $\mathbf{W}$, $\mathbf{V}$, $\mathbf{c}_l$.
\end{algorithmic}
\end{algorithm}
The advantages of this method compared to the state-of-the-art methods can be summarized as follows.
First, the proposed solution takes into account the structure of the Jacobian tensor, as opposed to the original work \cite{Dreesen2015}. Compared to the constrained solutions \cite{Hollander2018,Zniyed2021}, the proposed solution proposes a new projection strategy. It is worth noting that our method constrains the factor $\mathbf{H}$, in the same spirit as in \cite{Hollander2018}, but it involves the pseudo-inverse of a matrix of size $rN \times rd$, instead of a large matrix of size $Nmn \times rd$ in \cite{Hollander2018}. This has the advantage of having much reduced computational and storage costs. In \cite{Hollander2018}, the author does not use Step 5 in Algorithm \ref{algo_rec2}, and the coefficients are estimated as follows.
$$\hat{\mathbf{c}} = \min_{\mathbf{c}}  \left \|  {\rm vec}\big({\rm unfold}_3 \boldsymbol{\mathcal{J}}\big) - {\rm vec}\big(\mathbf{H} (\mathbf{V} \odot \mathbf{W})^T\big) \right \|^2.$$
In addition, the method in \cite{Zniyed2021} imposes the constraint on the rank-one tensors instead of the factors. This method \cite{Zniyed2021} has the advantage of having better convergence compared to \cite{Hollander2018}. Nevertheless, the method in \cite{Zniyed2021} is computationally heavier, compared to factor constraining solutions, since it uses iterated projections of tensors instead of factor matrices. Finally, the state-of-the-art methods \cite{Dreesen2015, Hollander2018, Zniyed2021} have been proposed, in nonlinear system identification community, for polynomial functions decoupling, whereas here, our method can handle different basis functions.

However, the CTD-based proposed approach, like the other state-of-the-art approaches, is based on the first-order information only. This has the disadvantage of not being able to estimate the constant terms of the approximated function. This will introduce a bias in the estimation. To overcome this problem, we propose in the next section to combine the first-order information with the zeroth-order information. This allows to solve the bias problem. Moreover, we will show in the sequel that we will have performances improvements when fusing these informations.

\section{Constrained CMTF-based approach}
\label{sec6}
In this section, we present the proposed CMTF-based solution for flexible NN learning. First, we propose a new formulation of the problem as a coupled matrix-tensor factorization problem \cite{Acar2011AllatonceOF,Boudehane2019}. Indeed, since the proposed solution is applied to pretrained NNs, we will take advantage of the zeroth-order information, by using a matrix $\mathbf{F}$, which contains the evaluations of the original pretrained subnetwork in $N$ different sampling points. We will show that this matrix shares a common factor with the Jacobian tensor $\boldsymbol{\mathcal{J}}$, thus a coupled decomposition is performed. The proposed approach fuses the first and zeroth order information, which allows to overcome the problems of the CTD-based approach, and to improve the performances. Once the problem is formulated, we will present the learning algorithm. This algorithm can be used with different bases as will be shown in what follows.

\subsection{Zeroth-order information}
Let $\mathbf{F}$ be a matrix of size $n \times N$ that contains the function evaluation at $N$ different sampling points $\mathbf{u}^{(j)} \in \mathbb{R}^m$. This matrix is defined as
\begin{align}
\label{matrix_F}
\mathbf{F} = \begin{bmatrix}
f_1 (\mathbf{u}^{(1)})& \cdots & f_1 (\mathbf{u}^{(N)}) \\ 
\vdots &  & \vdots \\ 
f_n (\mathbf{u}^{(1)}) & \cdots & f_n (\mathbf{u}^{(N)})
\end{bmatrix}.
\end{align}
Based on \eqref{matrix_F} and \eqref{eq_problem_equiv}, we can see that $\mathbf{F}$ is subject to a rank factorization given by 
\begin{align}
\label{matrix_F_expression}
\mathbf{F} = \begin{bmatrix}
f_1 (\mathbf{u}^{(1)})& \cdots & f_1 (\mathbf{u}^{(N)}) \\ 
\vdots &  & \vdots \\ 
f_n (\mathbf{u}^{(1)}) & \cdots & f_n (\mathbf{u}^{(N)})
\end{bmatrix} = \mathbf{W} \cdot \underbrace{\begin{bmatrix}
g_1 (\mathbf{v}_1^T \mathbf{u}^{(1)})& \cdots & g_1 (\mathbf{v}_r^T \mathbf{u}^{(N)}) \\ 
\vdots &  & \vdots \\ 
g_r (\mathbf{v}_1^T \mathbf{u}^{(1)}) & \cdots & g_r (\mathbf{v}_r^T \mathbf{u}^{(N)})
\end{bmatrix}}_{\mathbf{Z}^T},
\end{align}
where $\mathbf{Z}$ is of size $N \times r$. We can see from \eqref{matrix_F_expression}, that matrix $\mathbf{F}$ shares the factor $\mathbf{W}$ with tensor $\boldsymbol{\mathcal{J}}$. Based on this observation, we will therefore use matrix $\mathbf{F}$ together with tensor $\boldsymbol{\mathcal{J}}$ in a constrained CMTF problem to train the flexible NN.

\subsection{Problem statement}
Now that we have defined the matrix $\mathbf{F}$ and the tensor $\boldsymbol{\mathcal{J}}$, we propose to formulate the learning problem as a constrained CMTF problem where we decompose both $\boldsymbol{\mathcal{J}}$ and $\mathbf{F}$, while enforcing the AFs to follow the form \eqref{model_AF}.
This problem can be expressed by the following criterion:
\begin{align}
\label{eq_crit}
\min_{\mathbf{W},\mathbf{V},\mathbf{H}, \mathbf{Z}} \Big \|  \boldsymbol{\mathcal{J}} - [| \mathbf{W},\mathbf{V},\mathbf{H} |] \Big \|^2 + \lambda \cdot \Big \|  \mathbf{F} -  \mathbf{W} \mathbf{Z}^T \Big \|^2 \\
\mbox{s.t.}  ~\ ~\ \mathbf{h}_l = \mathbf{X}_l \cdot \mathbf{c}_l,  ~\ ~\  \mathbf{z}_l = \mathbf{Y}_l \cdot \mathbf{c}_l, ~\ ~\ ~\ \mbox{for} ~\ 1 \leq l \leq r \notag
\end{align}
where the matrices $\mathbf{X}_l$ and $\mathbf{Y}_l$, of size $N \times (d+1)$, are expressed as
\begin{align}
\label{matrix_X_l}
\mathbf{X}_l= \begin{bmatrix}
0 & \phi_1' (\mathbf{v}_l^T \mathbf{u}^{(1)})& \cdots & \phi_d' (\mathbf{v}_l^T \mathbf{u}^{(1)}) \\ 
\vdots & \vdots &  & \vdots \\ 
0 & \phi_1' (\mathbf{v}_l^T \mathbf{u}^{(N)}) & \cdots & \phi_d' (\mathbf{v}_l^T \mathbf{u}^{(N)})
\end{bmatrix},
\end{align}
\begin{align}
\label{matrix_Y_l}
\mathbf{Y}_l= \begin{bmatrix}
1 & \phi_1 (\mathbf{v}_l^T \mathbf{u}^{(1)})& \cdots & \phi_d (\mathbf{v}_l^T \mathbf{u}^{(1)}) \\ 
\vdots & \vdots &  & \vdots \\ 
1 & \phi_1 (\mathbf{v}_l^T \mathbf{u}^{(N)}) & \cdots & \phi_d (\mathbf{v}_l^T \mathbf{u}^{(N)})
\end{bmatrix}.
\end{align}
The constraints in \eqref{eq_crit} mean that the entries in matrices $\mathbf{H}$ and $\mathbf{Z}$ are expressed as
\begin{align}
    h_{j,l}=g_l'(\mathbf{v}_l^T \mathbf{u}^{(j)}), && z_{j,l}=g_l(\mathbf{v}_l^T \mathbf{u}^{(j)}).
\end{align}
Solving the above criterion allows to retrieve the weights and the AFs parameters from both $\boldsymbol{\mathcal{J}}$ and $\mathbf{F}$, while forcing the AFs to follow \eqref{model_AF}. It is worth noting that the use of matrix $\mathbf{F}$ allows to estimate the constant terms $c_{0,l}$, contained in matrix $\mathbf{Z}$, which is not the case with tensor $\boldsymbol{\mathcal{J}}$. Moreover, this fusion allows an improvement of performances, which will be shown later.

\subsection{Learning algorithm}
To solve the problem \eqref{eq_crit}, we propose an alternating least squares algorithm with constrained factors $\mathbf{H}$ and $\mathbf{Z}$. These constraints will allow to update the coefficients of the AFs $g_1, \cdots, g_r$ according to \eqref{model_AF}. 
The proposed algorithm works as follows. First, we estimate the factors one by one, by fixing all but the factor to be estimated and finding its optimal solution to a minimization problem. This first minimization does not take into account the structure of $\mathbf{H}$ and $\mathbf{Z}$.  Second, we find an optimal solution of the coefficients $\mathbf{c}_l$ which minimizes a cost function that considers both $\mathbf{H}$ and $\mathbf{Z}$. Finally, we project the columns $\mathbf{h}_l$ and $\mathbf{z}_l$ according to the constraints defined in \eqref{eq_crit}. The full procedure is given in Algorithm \ref{alg_learning}.

\begin{algorithm}
\caption{CMTF-based learning algorithm}
\label{alg_learning}
\begin{flushleft}
\textbf{Input:} Tensor $\boldsymbol{\mathcal{J}}$ of size $n \times m \times N$, matrix $\mathbf{F}$ of size $n \times N$, functions $\{\phi_1, \cdots, \phi_d\}$, fitting parameter $\lambda$, rank $r$\\
\textbf{Output:} Factors $\mathbf{W}$, $\mathbf{V}$, $\mathbf{H}$ and $\mathbf{Z}$, and coefficients $\mathbf{c}_l$.
\end{flushleft}
\begin{algorithmic}
\STATE Initialize $\mathbf{V}$, $\mathbf{H}$, $\mathbf{Z}$
\REPEAT
\STATE
\STATE \hrule
\STATE{Update $\mathbf{W}$ with $\displaystyle\min_{\mathbf{W}} \Big \| {\rm unfold}_1 \boldsymbol{\mathcal{J}} -  \mathbf{W} \Big( \mathbf{H} \odot \mathbf{V} \Big)^T \Big \|^2 + \lambda \cdot \Big \|  \mathbf{F} -  \mathbf{W} \mathbf{Z}^T \Big \|^2$}
\STATE{Update $\mathbf{V}$ with $\displaystyle\min_{\mathbf{V}} \Big \| {\rm unfold}_2 \boldsymbol{\mathcal{J}} -  \mathbf{V} \Big( \mathbf{H} \odot \mathbf{W} \Big)^T \Big \|^2$ }
\STATE{Update $\mathbf{H}$ with $\displaystyle\min_{\mathbf{H}} \left \| {\rm unfold}_3 \boldsymbol{\mathcal{J}} -  \mathbf{H} \Big( \mathbf{V} \odot \mathbf{W} \Big)^T \right \|^2$ }
\STATE{Update $\mathbf{Z}$ with $\displaystyle\min_{\mathbf{Z}} \left \| \mathbf{F} -  \mathbf{W} \mathbf{Z}^T \right \|^2$ }
\STATE
\STATE \hrule
\FOR{$l = 1 \cdots r$}
\STATE{Construct matrix $\mathbf{X}_l$ according to \eqref{matrix_X_l}.}
\STATE{Construct matrix $\mathbf{Y}_l$ according to \eqref{matrix_Y_l}.}
\ENDFOR
\STATE{Compute $\mathbf{c}$ as $\displaystyle\min_{\mathbf{c}}  \left \| {\rm vec}(\mathbf{H}) - \begin{bmatrix}
\mathbf{X}_1 &  & \text{\large0}\\
 &\ddots & \\
\text{\large0} & & \mathbf{X}_r
\end{bmatrix} \cdot \mathbf{c} \right \|^2 + \lambda \cdot \left \| {\rm vec}(\mathbf{Z}) - \begin{bmatrix}
\mathbf{Y}_1 &  & \text{\large0}\\
 &\ddots & \\
\text{\large0} & & \mathbf{Y}_r
\end{bmatrix} \cdot \mathbf{c} \right \|^2$}
\STATE
\STATE \hrule
\FOR{$l = 1 \cdots r$}
\STATE{Compute $\mathbf{h}_l$ such that: ~\ ~\ $\mathbf{h}_l = \mathbf{X}_l \cdot \mathbf{c}_l$}
\STATE{Compute $\mathbf{z}_l$ such that: ~\ ~\ $\mathbf{z}_l = \mathbf{Y}_l \cdot \mathbf{c}_l$}
\ENDFOR
\STATE
\STATE \hrule
\UNTIL{a convergence test is satisfied or maximum iterations exhausted}
\STATE
\STATE \hrule
\RETURN $\mathbf{W}$, $\mathbf{V}$, $\mathbf{c}_l$.
\end{algorithmic}
\end{algorithm}
The AFs coefficients are estimated at the same time in the vector $\mathbf{c}$. Vector $\mathbf{c}$ is of length $r(d+1)$, and is defined as $$\mathbf{c} = [\mathbf{c_1}; \cdots; \mathbf{c}_r].$$ 
The form of the update of  $\mathbf{c}$ comes from the vectorization of the following problem
\begin{align}
    \displaystyle\min_{\mathbf{c}}  \Big \| \mathbf{H} - [\mathbf{X}_1 \mathbf{c}_1, \cdots, \mathbf{X}_r \mathbf{c}_r] \Big \|^2 + \lambda \cdot \Big \| \mathbf{Z} - [\mathbf{Y}_1 \mathbf{c}_1, \cdots, \mathbf{Y}_r \mathbf{c}_r] \Big \|^2,
\end{align}
which leads to block-diagonal matrices as in Algorithm \ref{alg_learning}.
In the next subsection, we will give more details on the implementation of Algorithm \ref{alg_learning}.

\subsection{Factors update}
The form of the updates of factors $\mathbf{V}$, $\mathbf{H}$ and $\mathbf{Z}$ in Algorithm \ref{alg_learning} is commonly known since they involve linear least squares problems.
Regarding $\mathbf{W}$ and $\mathbf{c}$, they will have the following forms
\begin{align}
\label{upd_W}
\mathbf{\hat{W}}= \big[{\rm unfold}_1 \boldsymbol{\mathcal{J}}, \lambda \cdot \mathbf{F}\big] \cdot \big[ (\mathbf{H} \odot \mathbf{V})^T, \lambda \cdot \mathbf{Z}^T\big]^\dagger,
\end{align}
and
\begin{align}
\label{upd_c}
\mathbf{\hat{c}}= \left[\begin{bmatrix}
\mathbf{X}_1 &  & \text{\large0}\\
 &\ddots & \\
\text{\large0} & & \mathbf{X}_r
\end{bmatrix}; \lambda \cdot \begin{bmatrix}
\mathbf{Y}_1 &  & \text{\large0}\\
 &\ddots & \\
\text{\large0} & & \mathbf{Y}_r
\end{bmatrix} \right]^\dagger \cdot \left[{\rm vec}(\mathbf{H}); \lambda \cdot {\rm vec}(\mathbf{Z})\right]
\end{align}

From equations \eqref{upd_W} and \eqref{upd_c}, we can see that both solutions involve the pseudo-inverse of large matrices. A large matrix of size $r \times N(m+1)$ for $\mathbf{\hat{W}}$, and a large and sparse matrix of size $2rN \times r(d+1)$ for $\mathbf{\hat{c}}$.
For an efficient computation of the solutions, we propose to: $(i)$ consider sparse numerical representations of the block-diagonal matrices that allocate space for nonzero elements only for a better management of the storage space, and $(ii)$ use the special form of the pseudo-inverse of the Khatri-Rao product. Applying this to matrix $\mathbf{\hat{W}}$ allows to rewrite the solution in \eqref{upd_W} as
\begin{align}
\label{upd_W_opt}    
\mathbf{\hat{W}}= \big[{\rm unfold}_1 \boldsymbol{\mathcal{J}} \cdot (\mathbf{H} \odot \mathbf{V}) +  \lambda^2 \cdot \mathbf{F}\mathbf{Z}\big] \cdot \big[ (\mathbf{H}^T\mathbf{H}) * (\mathbf{V}^T\mathbf{V}) + \lambda^2 \cdot (\mathbf{Z}^T \mathbf{Z})\big]^\dagger.
\end{align}
The advantage of this version is that we calculate only the pseudo-inverse of an $r \times r$ matrix rather than a $r \times N(m+1)$ matrix. This formulation is also used for the other factors in the simulations.

\subsection{Toy example}
\begin{figure}[htbp]
\centering\includegraphics[scale=0.3]{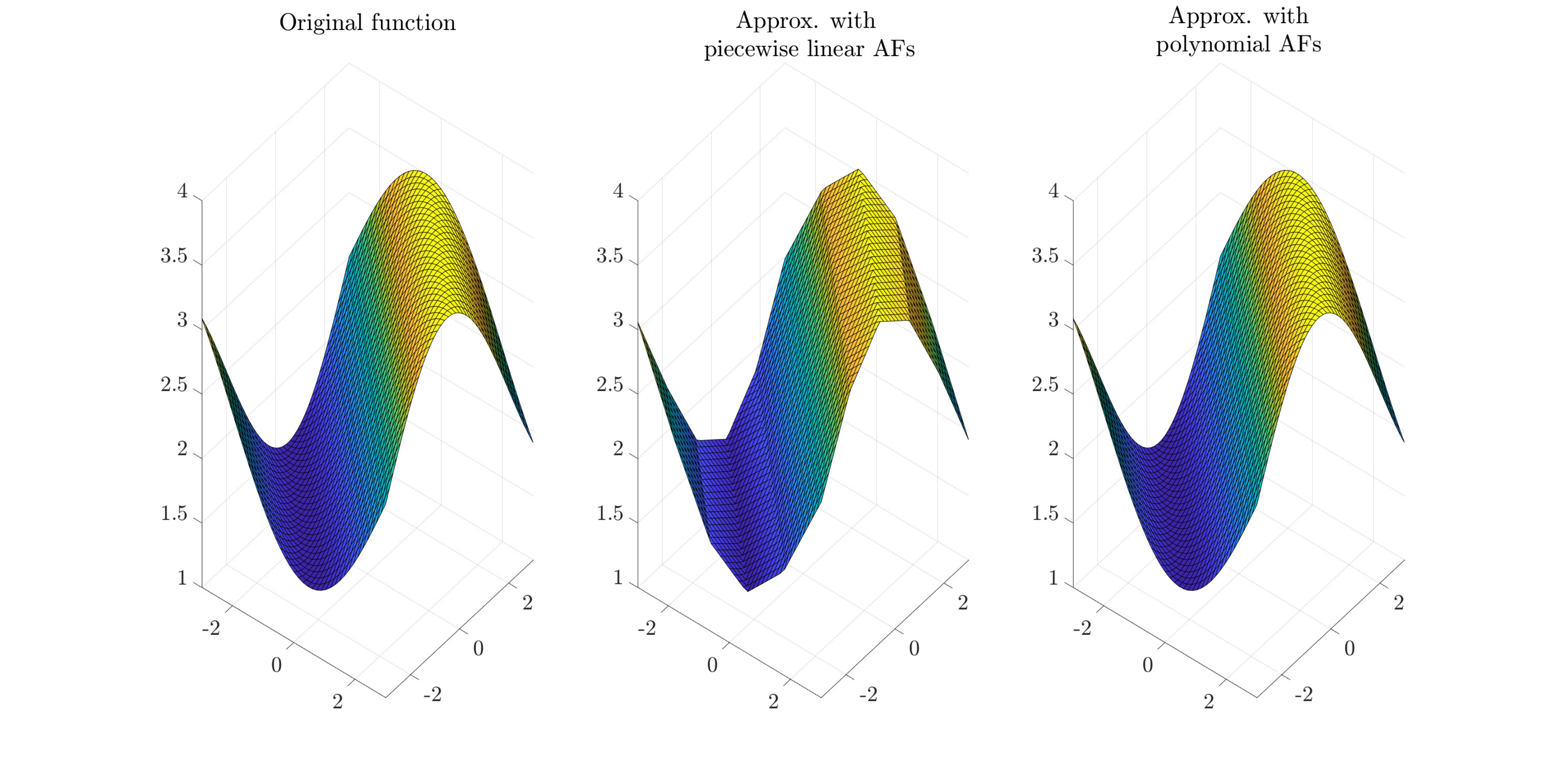}
\caption{The approximation of $f_1(u_1,u_2)=2.5+{\rm sin}\Big(0.2 \cdot \pi \cdot (u_1+u_2)\Big)$, with $r=3$, $d=10$.}
\label{fig_toy_example1}
\end{figure}

\begin{figure}[htbp]
\centering\includegraphics[scale=0.3]{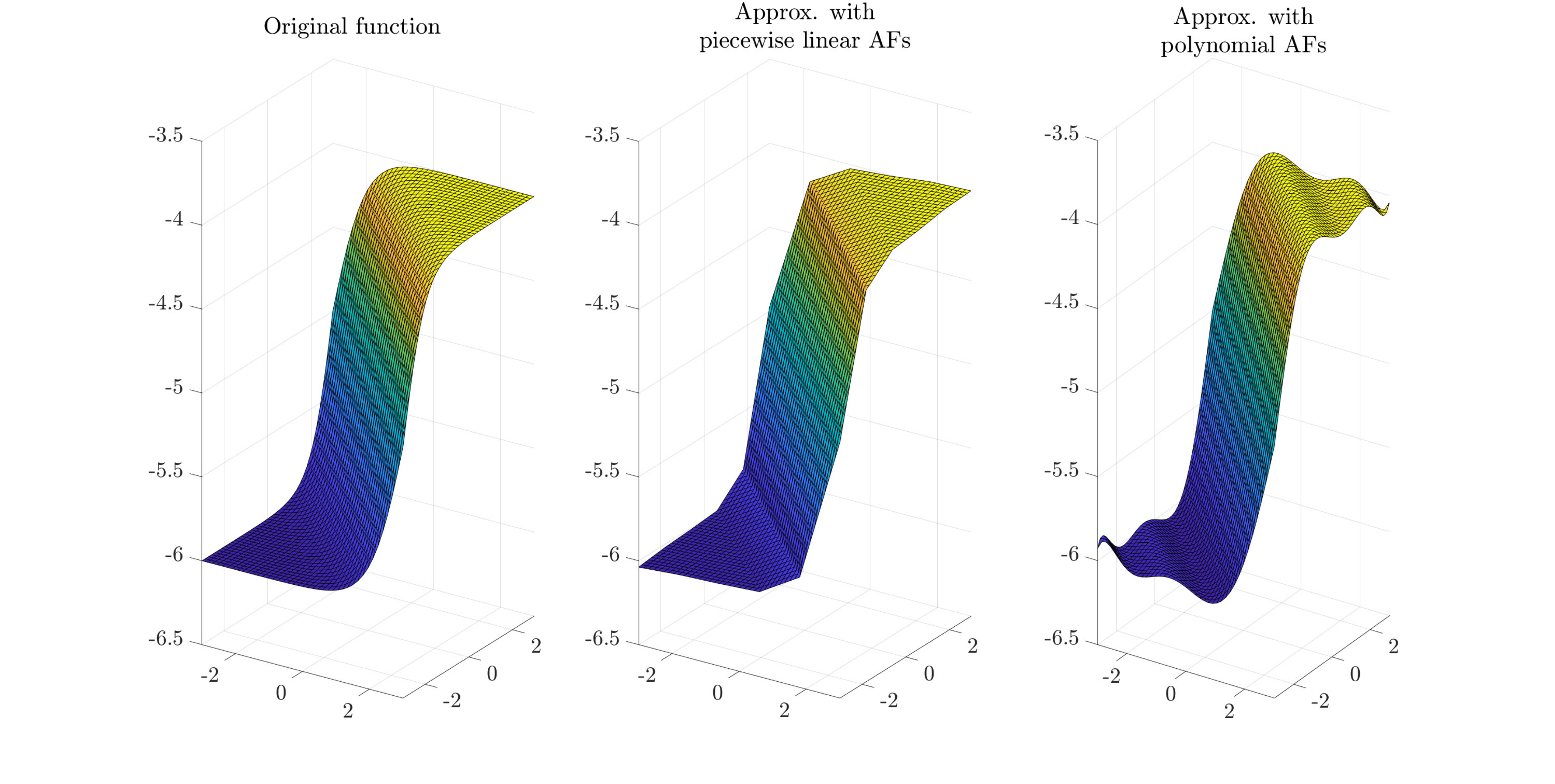}
\caption{The approximation of $f_2(u_1,u_2)=-5+{\rm tanh}\big(u_1+u_2\big)$, with $r=3$, $d=10$.}
\label{fig_toy_example2}
\end{figure}

Before applying the proposed model to NNs, we simulate, in this toy example, a multivariate function with $2$ inputs and $2$ outputs. We represent in the left of Figs. \ref{fig_toy_example1} and \ref{fig_toy_example2}, the original functions $f_1(u_1, u_2)$ and $f_2(u_1, u_2)$, respectively. Function $f_1(u_1, u_2)$ has a sinusoidal shape with respect to the sum of the inputs, and $f_2(u_1,u_2)$ is a hyperbolic tangent as a function of the sum of the inputs. Both output functions are not centered around zero. Piecewise linear functions and polynomial functions are used to approximate the function. 
One can note that the curves in the middle are piecewise linear, just like the considered flexible AFs. This basis suits better for $f_2$. 
The curves in the right have polynomial shapes, and this corresponds more to the shape of $f_1$. However, both decomposition bases give accurate approximations of the original functions. It is also worth mentioning that the functions are approximated without an offset, which show that the CMTF approach allows to estimate directly the constant terms of the approximated function.
\label{simu_toy_ex}

\section{Neural networks compression}
\label{sec7}
In this section, we explain how the proposed method is applied to NNs, and show its effectiveness. Several experiments are performed, while studying the impact of the different parameters. Four different bases are considered.

This section is organized as follows.
\begin{enumerate}
\item[-] In section \ref{sec_app}, we describe the compression process. We try to explain how our method is applied to pretrained neural networks.
\item[-] In section \ref{sec_CNN}, we recall the description of the CNN proposed in \cite{10.1007/978-3-319-10593-2_34}, which is used for character classification from the ICDAR dataset. Subsequently, this network is used to show the effectiveness of the proposed method.
\item[-] In section \ref{simus_res}, we evaluate the performances of the proposed methods with different bases in application to the described CNN. Our experiments will be devoted to the approximation of one and two layers of the original CNN with the flexible NN model.
\end{enumerate}

\subsection{Application to neural networks}
\label{sec_app}
\begin{figure}[htbp]
\centering\includegraphics[scale=0.5]{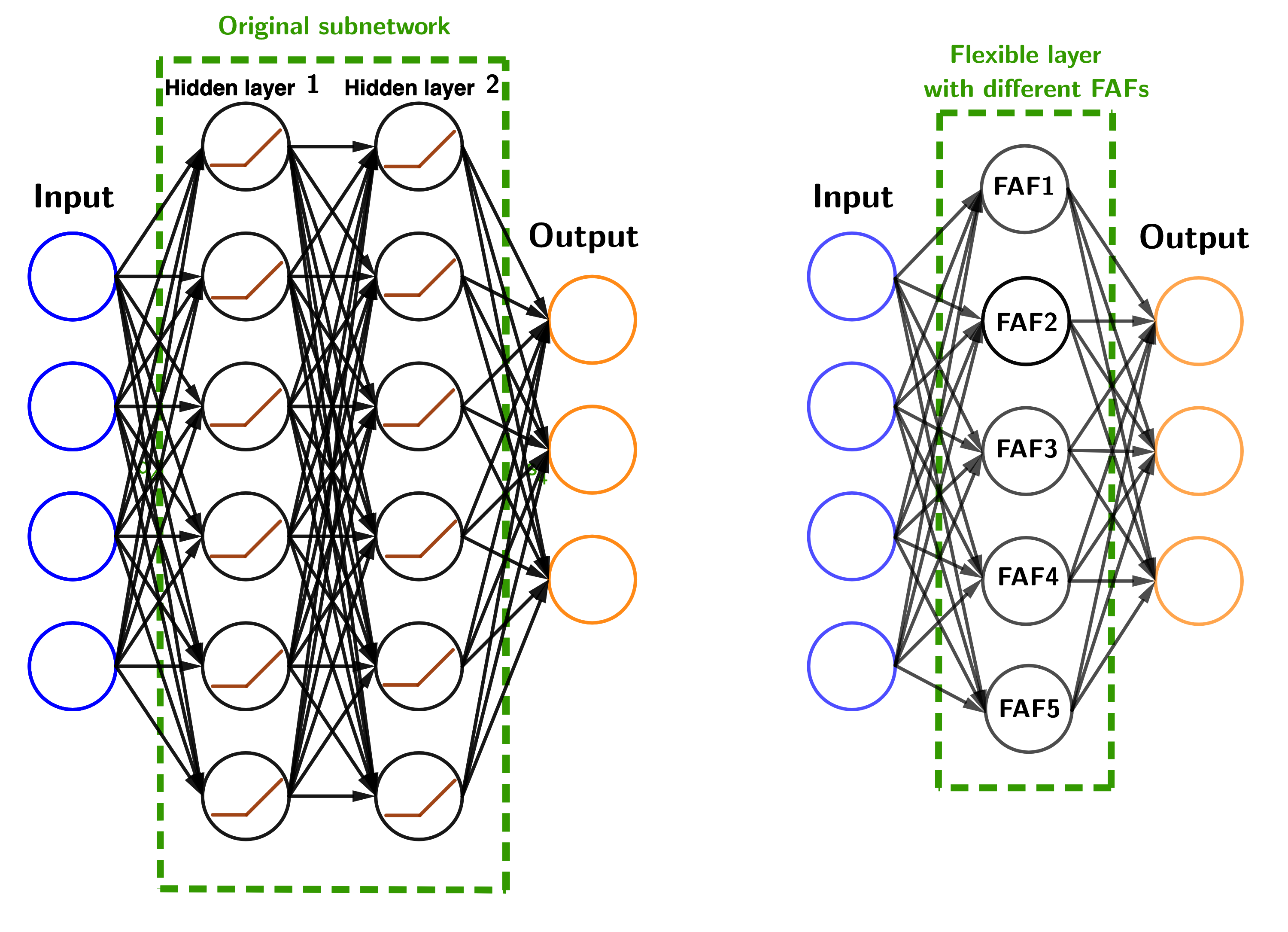}
\caption{Graphical representation of the compression process: (left) the original pretrained NN with fixed AFs - (right) the approximated NN with flexible AFs.}
\label{fig_comp}
\end{figure}

The idea of this work is to apply the model of the flexible NN presented in Section \ref{sec3} to large pretrained NNs. The goal is to replace one or more consecutive layers, namely the original subnetwork, by the flexible layer, whose function is described in \eqref{eq_problem_equiv}. This approximation/replacement can be applied to several parts of deep neural networks.
The aim being to approximate the input-output relationship of the original subnetwork with a multivariate function which can be seen as a flexible layer as presented in Fig. \ref{fig_fun}.
This process is represented graphically in Fig. \ref{fig_comp}, where we present on the left an example of a neural network with two hidden layers with fixed AFs, that are ReLUs in this example, and on the right the compressed network. The two hidden layers have been replaced by the proposed layer with flexible and different AFs for each neurons and with a different number of neurons in the hidden layer than the initial network.
The idea, in this example, is to consider a multivariate function defined as
$$\mathbf{f}(\mathbf{u}) = \mathbf{C} \cdot \mathbf{h}\Big(\mathbf{B}^T \mathbf{h}(\mathbf{A}^T \mathbf{u})\Big),$$
where $\mathbf{A}$, $\mathbf{B}$ and $\mathbf{C}$ are transformation matrices, and $\mathbf{h}$ is a vector function, whose elements are the ReLu function for example. This function is then approximated with a function $\hat{\mathbf{f}}(\mathbf{u})$ expressed as
$$\hat{\mathbf{f}}(\mathbf{u}) = \mathbf{W} \mathbf{g}(\mathbf{V}^T \mathbf{u}),$$
where the functions in vector $\mathbf{g}$ are all trainable and different.

By providing this flexibility in the network, the goal is to increase its expressiveness with the flexible AFs, in order to reduce the number of its parameters, while maintaining very good performance. The learning of the flexible layer is performed using the evaluation, on different sampling points, of the original subnetwork's function and its first-order derivative.

Regarding the basis functions in \eqref{model_AF}, different bases can be used. In this following, we have chosen to decompose the FAFs, without loss of generality, into $4$ formats, which are as follows.

\begin{enumerate}
    \item $g_l(t) = c_{0,l} + \sum_{k=1}^{d} c_{k,l} \cdot {\rm ReLU}(t-t_k)$, with $t_k =  \frac{k-1}{d} \cdot {\scriptstyle\max_l}$.
        \item $g_l(t) = c_{0,l} + \sum_{k=1}^{d} c_{k,l} \cdot {\rm ReLU}(t-t_k)$, with $t_k = {\scriptstyle\min_l} + \frac{k-1}{d} \cdot ({\scriptstyle\max_l} - {\scriptstyle\min_l})$.
    \item $g_l(t) = c_{0,l} + c_{1,l} \cdot {\rm ReLU}(-t) + \sum_{k=1}^{d-1} c_{k+1,l} \cdot {\rm ReLU}(t-t_k)$, with $t_k = \frac{k-1}{d-1} \cdot {\scriptstyle\max_l}$.
    \item $g_l(t) = \sum_{k=0}^{d} c_{k,l} \cdot t^k$.
\end{enumerate}
In the above functions, ${\scriptstyle\max_l}$ and ${\scriptstyle\min_l}$ represent the highest and lowest values of the input data $t$. The entries $t$ in the above functions are related to $\mathbf{v}_l^T \mathbf{u}^{(j)}$. This means that the highest and lowest values change with the matrix $\mathbf{V}$. Therefore, $\max_l$ and $\min_l$, for $1 \leq l \leq r$, must be calculated as follows.
\begin{align}
     {\scriptstyle\max_l} = \max_{1 \leq j \leq N} \big(\mathbf{v}_l^T \mathbf{u}^{(j)} \big), &&
     {\scriptstyle\min_l} = \min_{1 \leq j \leq N} \big(\mathbf{v}_l^T \mathbf{u}^{(j)} \big).
\end{align}

\subsection{Character classification CNN and the ICDAR dataset}
\label{sec_CNN}
In this section, we will detail the CNN that will be used in the rest of the simulations. The network described in \cite{10.1007/978-3-319-10593-2_34} is a CNN with four convolutional layers using maxout \cite{Goodfellow2013} as the nonlinear AF. The network is trained to classify $24 \times 24$ gray-scale character images into one of $36$ classes, corresponding to $10$ digits and $26$ letters. The description of each layer is as follows.
\begin{itemize}
\item In the first layer, the input images are convolved with $2$ different groups of $48$ filters of size $9 \times 9$ each. The $2$ groups of channels are then pooled with maxout, resulting in $48$ channels with maps of size $16 \times 16$. To sum up, the input $\boldsymbol{\mathcal{X}}^{(0)}$ of the first layer is of size $24 \times 24$, the output $\boldsymbol{\mathcal{X}}^{(1)}$ is of size $16 \times 16 \times 48$, the weight tensor $\boldsymbol{\mathcal{W}}_{s_1}^{(1)}$ is of size $9 \times 9 \times 1 \times 48$ and the bias $\boldsymbol{\mathcal{B}}_{s_1}^{(1)}$ is of size $16 \times 16 \times 48$, for $1 \leq s_1 \leq 2$.
\item The second layer convolves its input with $2$ groups of $64$ filters each. The size of the filters is $9 \times 9$. The input $\boldsymbol{\mathcal{X}}^{(1)}$ is then of size $16 \times 16 \times 48$, the output $\boldsymbol{\mathcal{X}}^{(2)}$of size $8 \times 8 \times 64$,the $4$-order weight tensor $\boldsymbol{\mathcal{W}}_{s_2}^{(2)}$ is of size $9 \times 9 \times 48 \times 64$ and the bias $\boldsymbol{\mathcal{B}}_{s_2}^{(2)}$ is of size $8 \times 8 \times 64$, for $1 \leq s_2 \leq 2$.
\item The third layer has an input $\boldsymbol{\mathcal{X}}^{(2)}$ of size $8 \times 8 \times 64$, an output $\boldsymbol{\mathcal{X}}^{(3)}$ of size $1 \times 1 \times 128$, and $4$ groups with $128$ filters of size $8 \times 8$. The $4$-order weight tensor $\boldsymbol{\mathcal{W}}_{s_3}^{(3)}$ is of size $8 \times 8 \times 64 \times 128$ and the bias $\boldsymbol{\mathcal{B}}_{s_3}^{(3)}$ is of size $1 \times 1 \times 128$, for $1 \leq s_3 \leq 4$.
\item The fourth layer has the following parameters. A vector input $\boldsymbol{\mathcal{X}}^{(3)}$ of length $128$, a vector output $\boldsymbol{\mathcal{X}}^{(4)}$ of length $37$, $4$ groups of filters $\boldsymbol{\mathcal{W}}_{s_4}^{(4)}$ of size $1 \times 1 \times 128 \times 37$ and the bias $\boldsymbol{\mathcal{B}}_{s_4}^{(4)}$ of size $1 \times 1 \times 37$, for $1 \leq s_4 \leq 4$.
\end{itemize}
\begin{remark}
The third layer of the CNN, described above, can be seen as a fully connected neural network with maxout activation function. Indeed, the convolution operation of the input $\boldsymbol{\mathcal{X}}^{(2)}$ of size $8 \times 8 \times 64$ with tensors $\boldsymbol{\mathcal{W}}_{s_3}^{(3)}$ of size $8 \times 8 \times 64 \times 128$ is equivalent to the matrix product of the reshaped version of the input $\boldsymbol{\mathcal{X}}^{(2)}$, called $\mathbf{x}^{(2)}$ of size $4096 \times 1$ and the reshaped version $\mathbf{W}_{s_3}^{(3)}$ of the weights of size $4096 \times 128$, resulting in a vector output $\boldsymbol{\mathcal{X}}^{(3)}$ of length $128$, where $\boldsymbol{\mathcal{X}}^{(3)} = \max\Big({\mathbf{W}_{s_3}^{(3)}}^T \mathbf{x}^{(2)} + \boldsymbol{\mathcal{B}}_{s_3}^{(3)} \Big)$. Same reasoning can be applied for the last layer. Based on this remark, we can see that the proposed method can be applied on layers $3$ and $4$ of the described CNN.
\end{remark}
\begin{remark}
The number of parameters in the third layer represents almost $75\%$ of the total number of parameters of the trained CNN, with over $2$ million parameters, which makes this part a good target for applying a parameter reduction method.
\end{remark}
The CNN described above is trained on the ICDAR dataset. ICDAR is a dataset containing $24 \times 24$ character images, $10$ digits and $26$ letters. The images are normalized with zero mean and unit variance. This dataset has $185639$ images in the training set and $5198$ in the test set. The reported recognition accuracy \cite{10.1007/978-3-319-10593-2_34} of the CNN is $96.97\%$ on the training set and $91.52\%$ on the test set.

\subsection{Simulation results}
\label{simus_res}
The aim of the following experiments is threefold. First, we show the interest of using a coupled matrix-tensor decomposition, instead of considering a constrained decomposition of the Jacobian tensor only. Second, we apply the proposed CMTF-based method for compressing a single layer of CNN, namely the third layer. Finally, we perform simulations for the compression of two layers, the third and the fourth.

In the following, the Jacobian tensor is built according to the definition in Section \ref{sec4}. The Jacobian matrices are estimated using the AD techniques. The AD allows to have derivative evaluations at machine precision. See \cite{MatlabLink, Baydin2017} for more details.
The Normalized MSEs are defined according to $$\text{Tensor NMSE}= \frac{|| \hat{\boldsymbol{\mathcal{J}}} - \boldsymbol{\mathcal{J}} ||_F^2}{|| \boldsymbol{\mathcal{J}}||_F^2},$$ and, $$\text{Matrix NMSE}= \frac{|| \hat{\mathbf{F}} - \mathbf{F} ||_F^2}{|| \mathbf{F}||_F^2},$$ where $\hat{\boldsymbol{\mathcal{J}}}$ and $\hat{\mathbf{F}}$ refer to the estimated tensor and matrix, respectively. The \texttt{Accuracy drop} is defined as the accuracy difference between the original and the approximated CNN. The \texttt{Compression ratio} is equal to the ratio between the number of parameters of the approximated and the original layers. 

The parameter $\lambda$ in Algorithm \ref{alg_learning} is an adaptive parameter, in the sense that it starts with a value $\lambda_0$ for the first iteration, and it evolves over the iterations. In the following simulations, we choose $\lambda_0 = 10^{-3}$, and we multiply $\lambda$ with $\sqrt{10}$ after every $10$ iterations. This allows to give more weight to the matrix part of the criterion \eqref{eq_crit} over the iterations.
From the different iterations, we choose the solution that corresponds to the lowest value of the Matrix NMSE.
The factors in Algorithm \ref{alg_learning} are initialized with a relaxed ALS solution.

\subsubsection{CMTF vs Constrained Jacobian decomposition}
\begin{figure}[htbp]
\centering\includegraphics[scale=0.3]{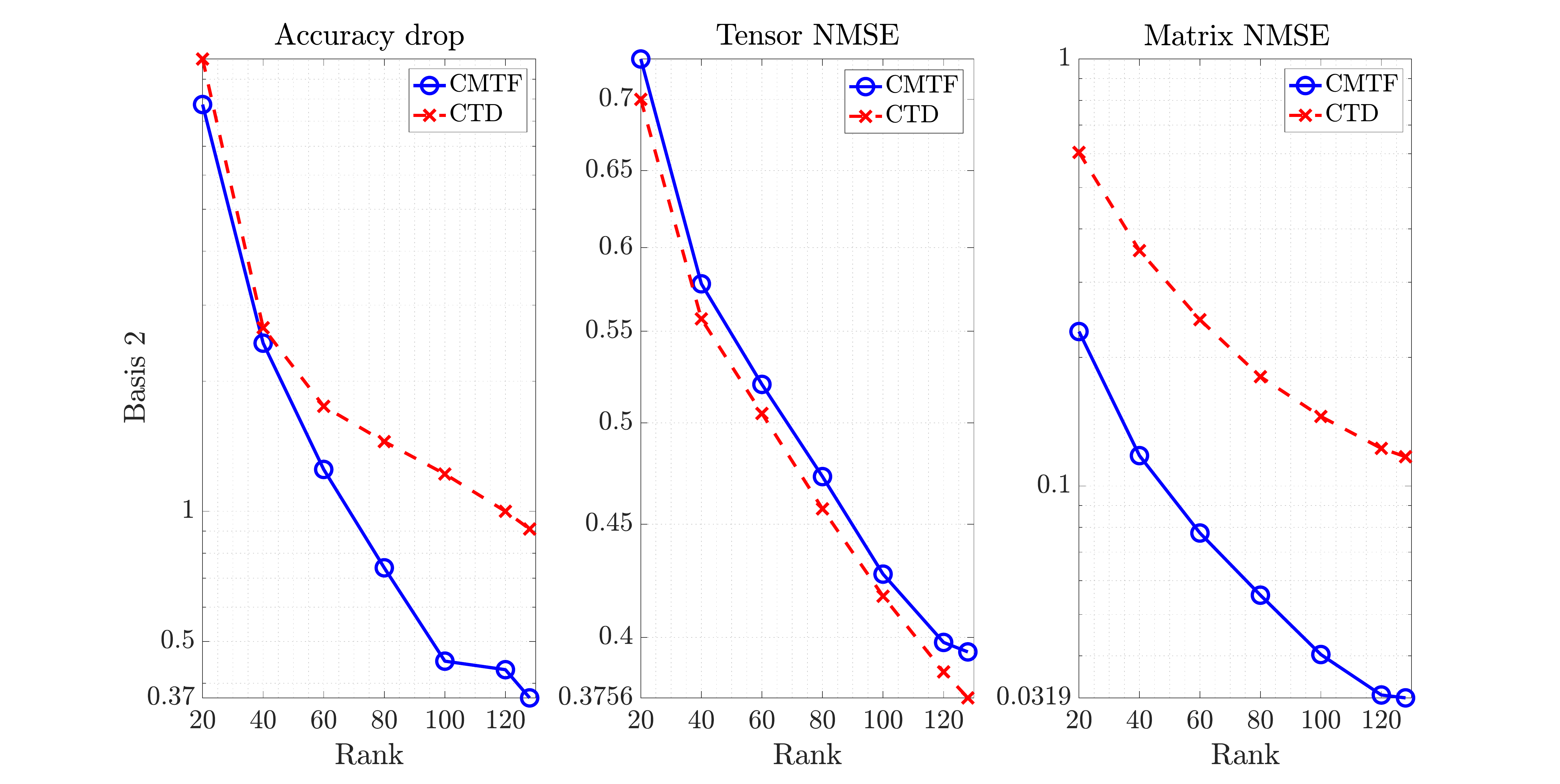}
\caption{CMTF vs CTD, with $N=360$, $d=4$.}
\label{fig_cmtf_vs_ctd}
\end{figure}

In Fig. \ref{fig_cmtf_vs_ctd}, we compare the proposed CMTF method with a constrained tensor decomposition of the Jacobian. In this experiment, we apply both methods to compress a single layer, namely the third one. Both methods use FAFs according to the third basis in Section \ref{sec_app}.
Since the CTD approach does not allow to estimate the constant terms, we perform an offset correction such that
$$\mathbf{\hat{f}}(\mathbf{u}) = \mathbf{W} \mathbf{g}(\mathbf{V}^T \mathbf{u}) + \Big(\mathbf{f}(\mathbf{0}) - \mathbf{\hat{f}}(\mathbf{0})\Big),$$
where $\mathbf{f}$ is the function of the original subnetwork, and $\mathbf{\hat{f}}$ is its approximation.
In this experiment, we fix $N=360$ and $d=4$. The Jacobian tensor $\boldsymbol{\mathcal{J}}$ is then of size $128 \times 4096 \times 360$ and the evaluation matrix $\mathbf{F}$ is of size $128 \times 360$.
Based on Fig. \ref{fig_cmtf_vs_ctd}, one can note that the CMTF approach gives a better classification accuracy compared to the CTD using the same decomposition basis. 
Regarding the tensor NMSE, we can remark a slight advantage of the CTD approach over the CMTF, while this advantage is more pronounced in favor of the CMTF in the matrix/function NMSE. 
Similar behavior has been found considering other decomposition bases.
This therefore shows the interest of using the coupled approach. First, to directly estimate the constant terms of the approximated function, and second, to have better performances. In the rest of the simulations, the learning is performed using the CMTF-based approach.

\subsubsection{Compression of a single layer}
\label{simus_single_layer} 
\begin{figure}[htbp]
\centering\includegraphics[width=\linewidth,height=12cm]{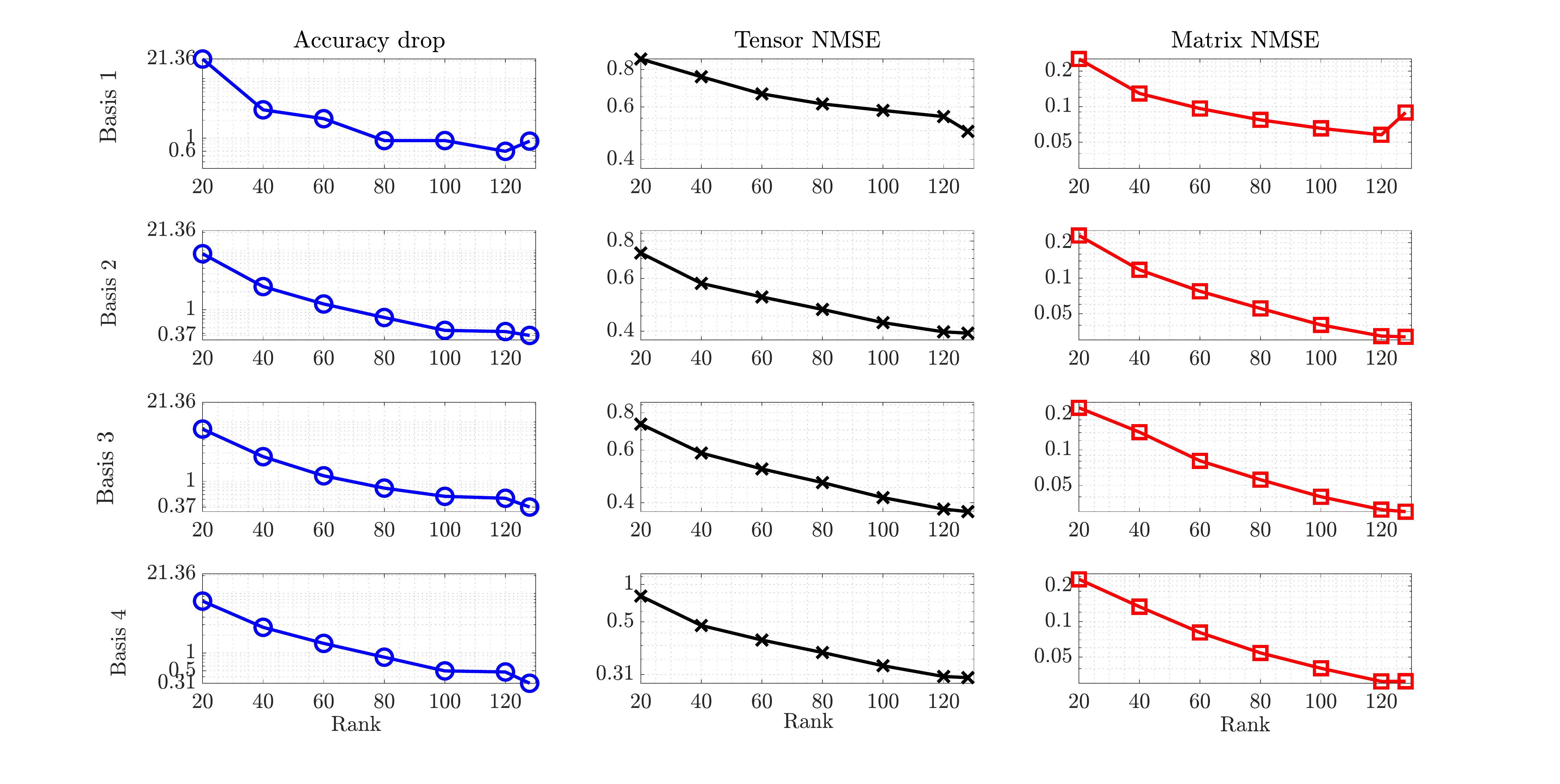}
\caption{Performances vs rank $r$, with $N=360$ and $d=4$.}
\label{fig_acc_rank}
\end{figure}

In this part, we compress the third layer of the CNN using the different decomposition bases. The impact of, both, the rank $r$ and the order $d$ are studied.
\paragraph{Impact of $r$}
In Fig. \ref{fig_acc_rank}, we plot the accuracy drop, the tensor and matrix NMSEs for the different bases. In this experiment, we vary the rank $r$, and we fix $N=360$ and $d=4$.
Different remarks can be drawn based on Fig. \ref{fig_acc_rank}. The different decomposition bases have an accuracy drop of less than $1\%$, compared to $91.52\%$ of the original network, starting from $r=80$. Bases $2$ and $4$ have an accuracy drop of less than $0.5\%$ from $r=100$, and basis $3$, from $r=120$. These $3$ bases, namely $2$ to $4$, are more flexible than base $1$ in the sense that their AFs allow to parameterize the negative part of their arguments, meanwhile in basis $1$, this part is always equal to zero.

\begin{figure}[htbp]
\centering\includegraphics[scale=0.25]{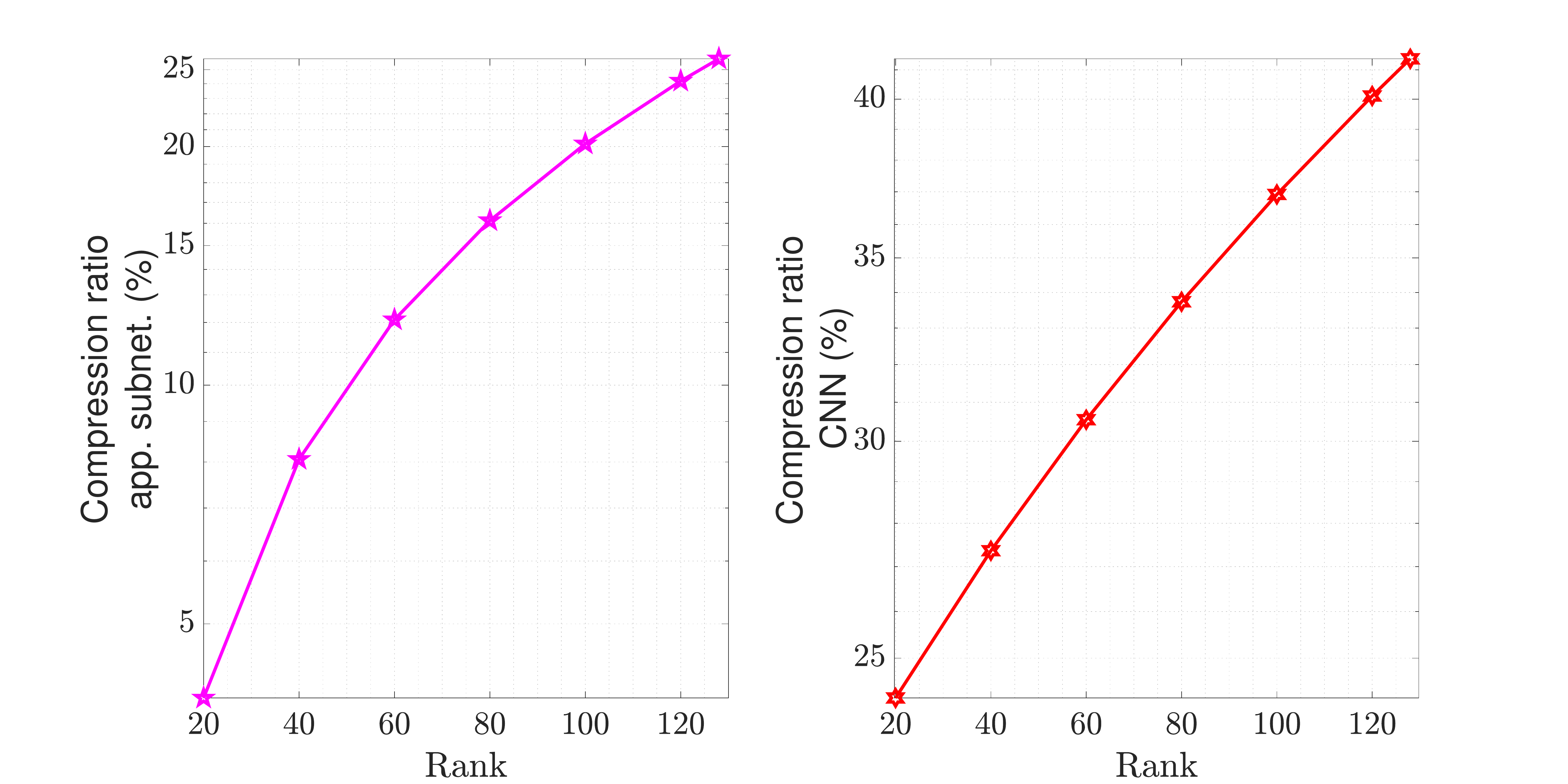}
\caption{Compression ratio vs rank $r$, with $N=360$ and $d=4$.}
\label{fig_comp_rank}
\end{figure}

\begin{figure}[htbp]
\centering\includegraphics[scale=0.25]{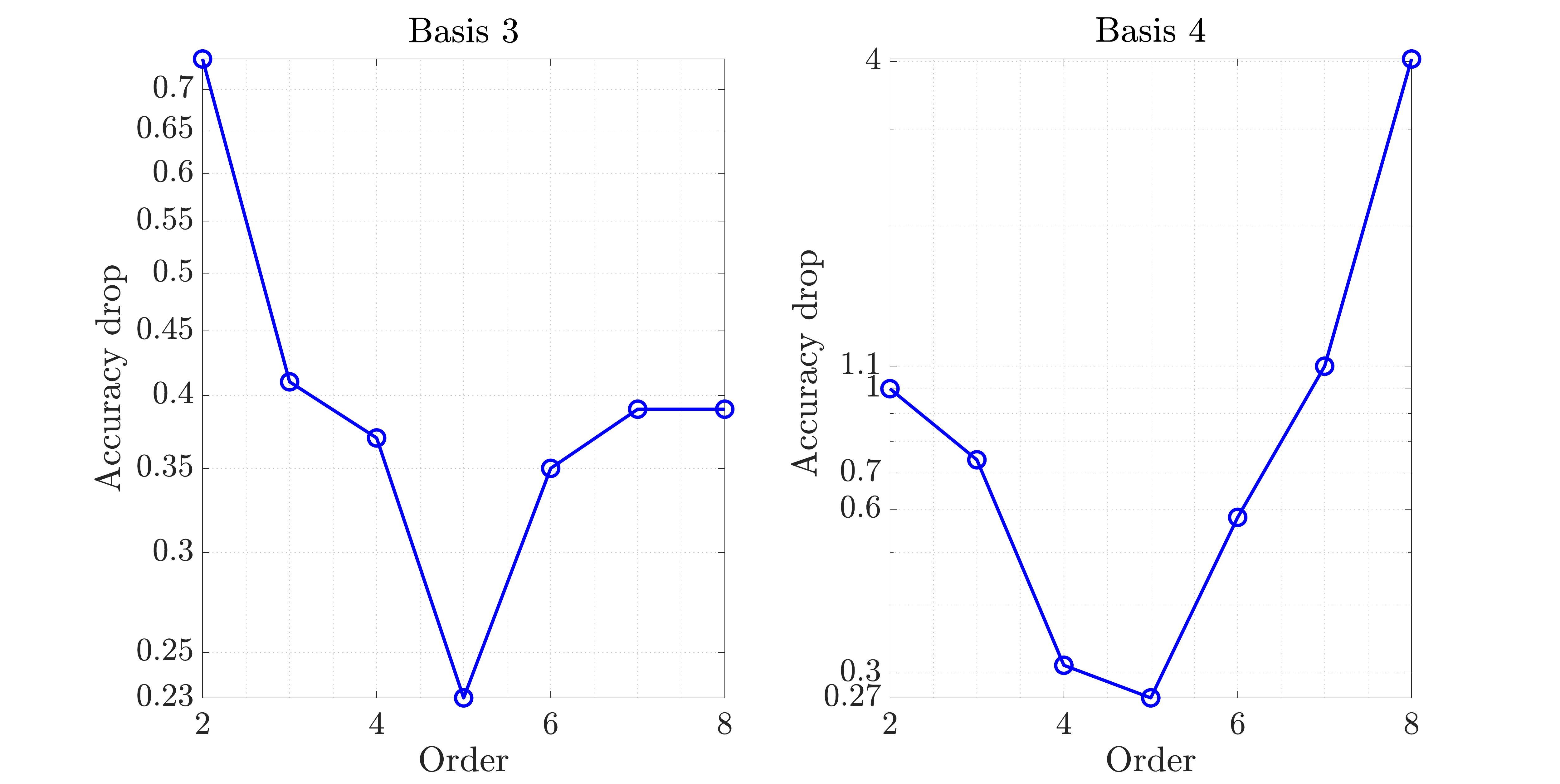}
\caption{Performances vs order $d$, with $N=360$ and $r=128$.}
\label{fig_acc_order}
\end{figure}

Regarding the compression ratio, the different bases have the same number of parameters for a fixed $r$ and $d$. We give in Fig. \ref{fig_comp_rank}, the compression ratio of the subnetwork and the compression ratio of the whole network with respect to the rank. 
This figure is to be analyzed at the same time with Fig \ref{fig_acc_rank}. We can notice that we have an accuracy drop of less than $0.5\%$ at $r=120$ for bases $2$ to $4$, while keeping only $25\%$ of the parameters of the original subnetwork in the flexible new layer, which corresponds to almost $40\%$ of the parameters of the whole original network that are kept in the new network. The best accuracy for $N=360$ and $d=4$ is $91.21\%$ for basis $4$, {\em i.e.,} an accuracy drop of $0.31\%$, corresponding to a compression ratio of almost $25\%$ on the subnetwork and almost $41\%$ on the whole network.
\paragraph{Impact of $d$}
In this section, we vary the order $d$, while fixing $N=360$ and $r=128$. Fig. \ref{fig_acc_order} gives the performance results with respect to the rank using bases $3$ and $4$.
The best accuracy is $91.29\%$, with basis $3$, $d=5$ and $r=128$, compared to $91.52\%$ of the original network. In this figure, we can see a phenomenon of over-fitting when the order $d$ is too high.

\begin{figure}[htbp]
\centering\includegraphics[scale=0.25]{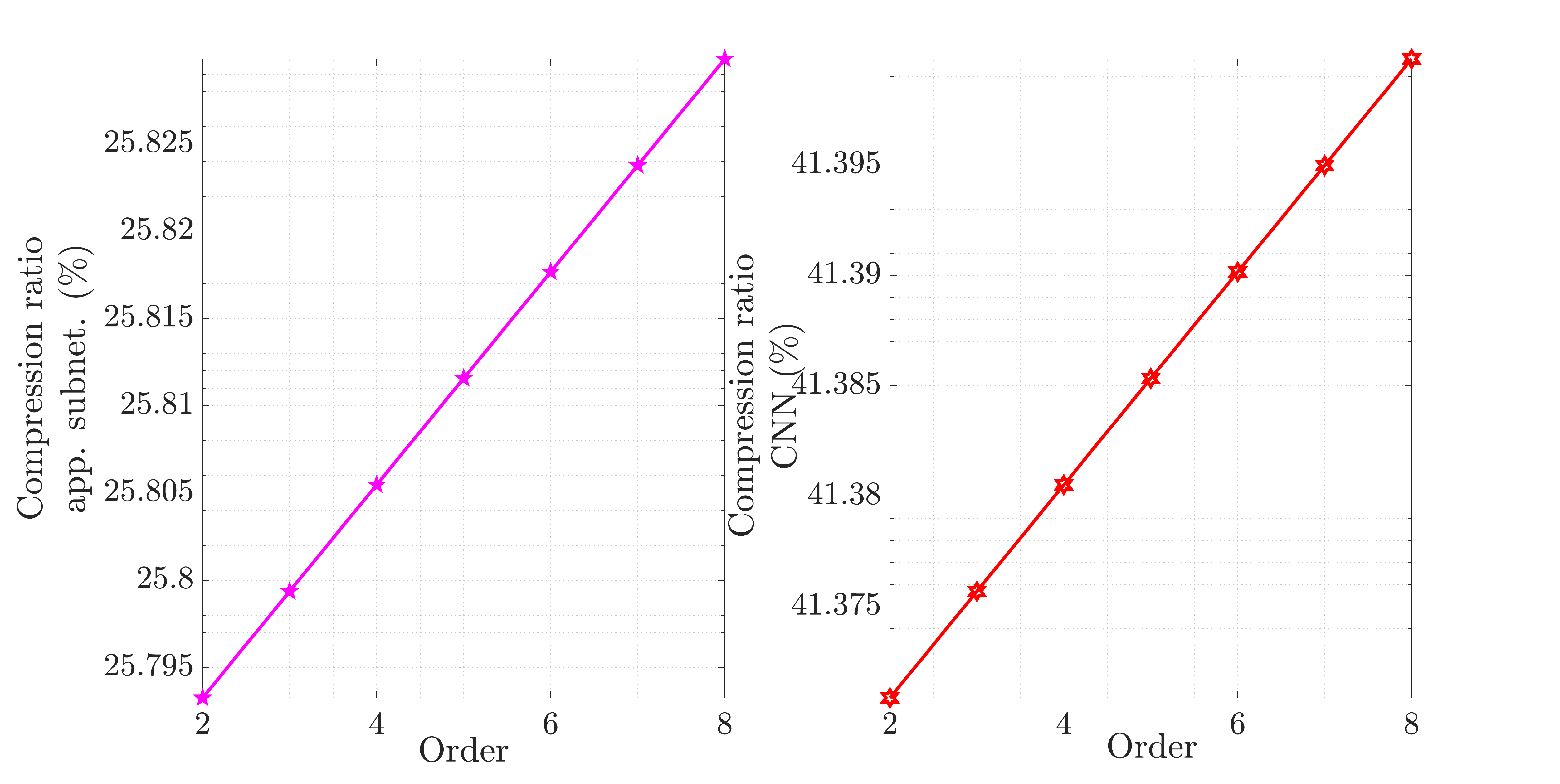}
\caption{Compression ratio vs order $d$, with $N=360$ and $r=128$.}
\label{fig_comp_order}
\end{figure}

In Fig. \ref{fig_comp_order}, we plot the compression ratios with respect to the order. We can notice that the parameter $d$ has a very low impact on the compression ratio when $r$ is fixed. The number of parameters in the flexible layer in this example is $4096 \cdot r+(d+1) \cdot r+128 \cdot r$.

\subsubsection{Compression of two layers}
\label{simus_two_layers}
\begin{figure}[htbp]
\centering\includegraphics[scale=0.3]{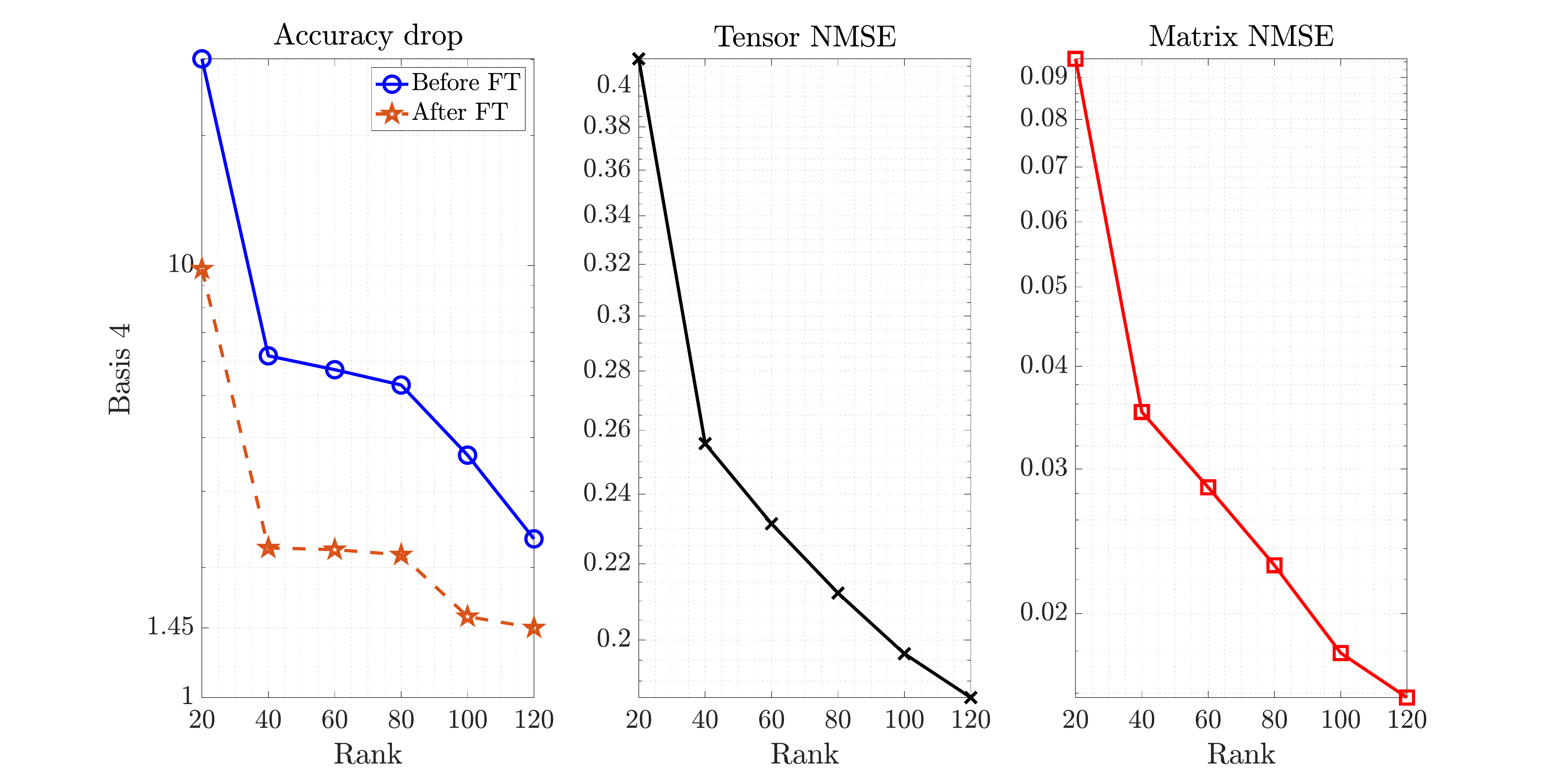}
\caption{Performances for two layers vs rank $r$, with $N=360$ and $d=4$.}
\label{fig_accuracy_two_layers}
\end{figure}

In this section, we replace two layers, the third and the fourth, of the original CNN with the flexible layer. The aim here is to show that the proposed solution works as well for compressing a single as several layers with the flexible layer. In the following experiment, we consider basis $4$, with $N=360$ and $d=4$.
In Fig. \ref{fig_accuracy_two_layers}, we plot the accuracy drop, the tensor and matrix NMSEs. One can note that we have higher accuracy drops compared to Section \ref{simus_single_layer}. This is because we compress two layers by one. To improve the performance, we perform a fine-tuning of the flexible layer. We can notice that we reach an accuracy drop of $1.45\%$, while keeping a lower number of parameters and with one layer less than the original network. This compression has two main advantages: $(i)$ reduce the number of parameters, and $(ii)$ speed up the network. 

\begin{figure}[htbp]
\centering\includegraphics[scale=0.25]{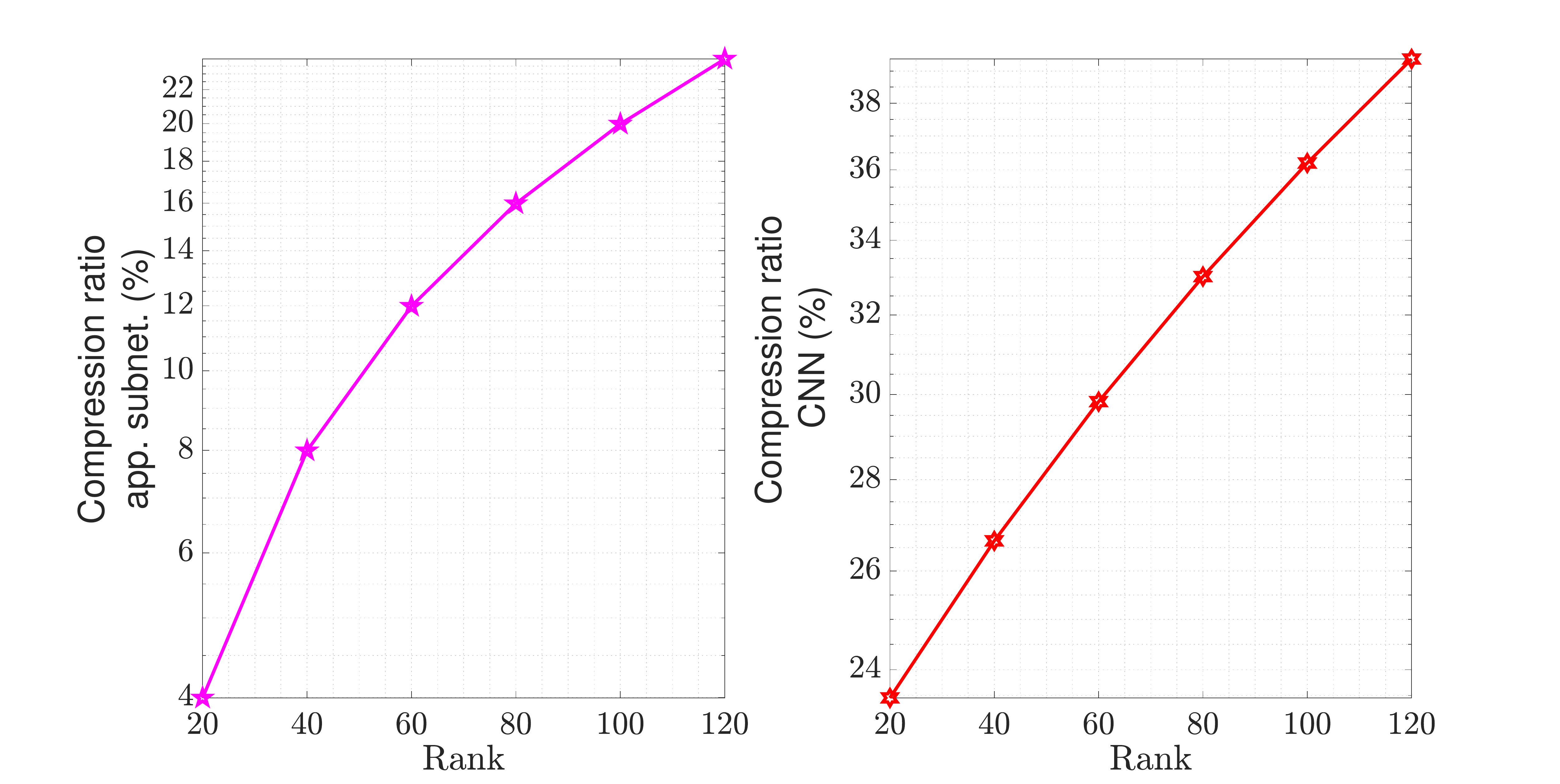}
\caption{Compression ratio for two layers vs rank $r$, with $N=360$ and $d=4$.}
\label{fig_compression_two_layers}
\end{figure}

Fig. \ref{fig_compression_two_layers} gives the compression ratios for this experiment. Overall, we can remark that we can reach an accuracy drop of less than $1.5\%$, while keeping less than $40\%$ of the parameters of the original network, and one less layer.

\section{Conclusion}
\label{sec8}
In this work, we have proposed a tensor-based framework for the compression of large neural networks, using flexible activation functions. The AFs are expressed as a weighted sum of predefined basis functions. By formulating the problem as a constrained coupled matrix-tensor factorization problem, we propose a new ALS-based learning algorithm, which can handle different decomposition bases. The aim of this solution is to compress large NN, by replacing one or more layers of the initial network with a layer having flexible AFs. The learning of the new functions is performed based on a coupled factorization of a Jacobian tensor, following a constrained canonical polyadic decomposition, and a constrained evaluation matrix. Several bases have been considered in this work. The efficiency of the proposed solution is shown in the simulations with the compression of subnetworks with one and two layers. Simulation results have shown that very satisfying compression ratios can be achieved with very low accuracy drops. In an example case, we have reached an accuracy drop of less than $0.25\%$, while keeping only $41.5\%$ of the parameters of the original network.

Another aim of this paper is to put forward the tensorial interpretation of neural networks. These tools have already been used for the compression of neural networks, but their use was usually reduced to finding low rank approximations of the weight tensors, without considering the improvement of the activation functions and the learning algorithms. We believe that this paper would trigger more research in these directions, from a tensor-based perspective. 

Perspective for future works include the modeling of several flexible layers using tensor-based methods, the generalization of the proposed approach to convolutional NNs, besides the training of FAFs directly from the data and not from the Jacobian tensor.
Identifiability of the considered decomposition in the case of piecewise linear functions deserves also to be investigated.

\bibliographystyle{siamplain}
\bibliography{references}

\end{document}